\documentclass[manuscript, screen, nonacm]{jair}

\JAIRTrack{}
\JAIRAE{}
\AtBeginDocument{%
  \fancyfoot{}%
  \fancyfoot[C]{\footnotesize\thepage}%
  \fancypagestyle{firstpagestyle}{\fancyhf{}\fancyfoot[C]{\footnotesize\thepage}}%
}

\RequirePackage[
  datamodel=acmdatamodel,
  style=acmauthoryear,
  backend=biber,
  giveninits=true,
  uniquename=init
  ]{biblatex}

\usepackage{tikz}
\usepackage{pgfplots}
\pgfplotsset{compat=1.18}
\usepackage{xcolor}
\usepackage{capt-of}
\definecolor{semcol}{HTML}{4CAF50}
\definecolor{themcol}{HTML}{FF9800}
\definecolor{salcol}{HTML}{1976D2}
\newcommand{\sub}{\hspace*{1.2em}}
\newlength{\taglen}
\newcommand{\trow}[3]{\par\noindent\hangindent=\taglen\hangafter=0
  \ifx\relax#2\relax\makebox[\taglen][l]{\textbf{[#1]}}#3\else
  {\color{#2}\makebox[\taglen][l]{\textbf{[#1]}}#3}\fi\par\vspace{2pt}}

\begin{document}

%%
%% The "title" command has an optional parameter,
%% allowing the author to define a "short title" to be used in page headers.
%\title{JAIR Example Template}
\title[Hierarchical Compositionality for Disambiguation]{
Hierarchical Compositionality for An Assistive AI Agent
}

%%
%% The "author" command and its associated commands are used to define
%% the authors and their affiliations.
%% Of note is the shared affiliation of the first two authors, and the
%% "authornote" and "authornotemark" commands
%% used to denote shared contribution to the research and/or corresponding author.
\author{Tianyi Fu}
\authornote{Corresponding Author.}
\orcid{0009-0007-3025-6778}
\email{Tianyi.Fu@ed.ac.uk}
\affiliation{%
  \institution{University of Edinburgh}
  \city{Edinburgh}
  \country{United Kingdom}
}

\author{Mohan Sridharan}
 \orcid{0000-0001-9922-8969}
\authornotemark[1]    
\email{m.sridharan@ed.ac.uk}
\affiliation{%
  \institution{University of Edinburgh}
  \city{Edinburgh}
  \country{United Kingdom}
}

\renewcommand{\shortauthors}{Fu \& Sridharan}
% \author{Lin Xu}
% \authornote{Corresponding Author.}
% \orcid{0000-0002-2037-3694}
% \email{xulin730@cs.ubc.ca}
% \affiliation{%
%   \institution{University of British Columbia}
%   \city{Vancouver}
%   \state{British Columbia}
%   \country{Canada}
% }

% \author{Frank Hutter}
% \orcid{0000-0002-2037-3694}
% \email{hutter@cs.ubc.ca}
% \affiliation{%
%   \institution{University of British Columbia}
%   \city{Vancouver}
%   \state{British Columbia}
%   \country{Canada}}

% \author{Holger H. Hoos}
% \orcid{0000-0003-0629-0099}
% \email{hoos@cs.ubc.ca}
% \affiliation{%
%   \institution{University of British Columbia}
%   \city{Vancouver}
%   \state{British Columbia}
%   \country{Canada}
% }

% \author{Kevin Leyton-Brown}
% \orcid{0000-0002-7644-5327}
% \email{kevinlb@cs.ubc.ca}
% \affiliation{%
%   \institution{University of British Columbia}
%   \city{Vancouver}
%   \state{British Columbia}
%   \country{Canada}
% }

% %% The short list of authors must be made of the list of all authors' lastnames.
% \renewcommand{\shortauthors}{Xu, Hutter, Hoo \& Leyton-Brown}
%% If this is too long and overlaps other information printed in the page headers, use
%\renewcommand{\shortauthors}{Xu et al.}

%%
%% The abstract is a short summary of the work to be presented in the
%% article.
\begin{abstract}
AI agents are increasingly being developed to assist humans in various applications, and Large Language Models and other deep network architectures are considered to be state of the art for such agents. These methods are impressive stochastic predictors, but they are resource-hungry, opaque, and known to make arbitrary decisions in novel situations due to the narrow set of underlying representation and processing choices. Our work seeks to explore the design of architectures for such AI agents based on core principles that can be traced back to the early pioneers of AI but are not fully utilized in modern AI methods. We do so in this paper in the context of the core problem of AI agents addressing ambiguity in the objects being referred to by the human participants. Humans address such ambiguity by heuristically leveraging compositional knowledge of domain context and the preferences of the other human participants. Drawing inspiration from this observation, we describe an architecture that embeds the principle of hierarchical compositionality and uses simple heuristics to achieve the desired disambiguation. 
% Specifically, domain objects are represented in terms of primitive attributes identified using a Large Language Model, and a hierarchical combination of attributes and \textit{concepts} automatically identified from a limited observed history of interactions of an assistive agent with specific users. 
Specifically, domain objects are represented in terms of primitive attributes drawn from human-validated semantic feature norms, and a hierarchical combination of \textit{attributes} and \textit{concepts} automatically identified from a limited observed history of interactions of an assistive agent with specific users. The assistive agent then achieves the desired disambiguation by reasoning with knowledge  of this compositional hierarchy; axioms governing domain dynamics; and models of semantic compatibility, session salience, and user-specific thematic preference, requesting human clarification when necessary. Experiments show that our approach consistently outperforms state of the art data-driven baselines, supporting adaptation to specific user profiles. Project website: \textcolor{red}{\url{https://tianyi-fu.github.io/HCAA/}}
\end{abstract}

%\begin{abstract}
%      A clear and well-documented \LaTeX\ document is presented as an
%  article formatted for publication by ACM in a conference proceedings
%  or journal publication. Based on the ``acmart'' document class, this
%  article presents and explains many of the common variations, as well
%  as many of the formatting elements an author may use in the
%  preparation of the documentation of their work.
%\end{abstract}

%% JAIR Note: 
%% Do not include ACM CCS Concepts or Keywords

% %% To be updated by authors.
% \received{20 February 2007}
% \received[accepted]{5 June 2009}

%%
%% This command processes the author and affiliation and title
%% information and builds the first part of the formatted document.
\maketitle

%%%%%%%%%%%%%%%%%%%%%%%%%%%%%%%%%%%%%%%%%%%%%%%%%%%%%%%%%%%%%%%%%%%
%%%%%%%%%%%%%%%%%%%%%%%%%%%%%%%%%%%%%%%%%%%%%%%%%%%%%%%%%%%%%%%%%%%
\section{Motivation}
\label{sec:introduction}
We are witnessing a rapid proliferation of AI agents being deployed to assist humans in various applications. Deep networks and \textit{foundation models} (FMs) such as Large Language Models (LLMs), Vision Language Models (VLMs), and Vision Language Action Models (VLAs) are considered state of the art for the design of such AI agents. These methods have been reported to provide impressive performance in perception, reasoning, and interaction problems, with the underlying stochastic prediction system elegantly capturing the most likely patterns in the training data~\cite{black:rss25,brohan:corl23,doshi:corl24,huang2022language,liu2023llm+}. At the same time, these methods are opaque \textit{black box} systems that are resource-hungry, requiring considerable training examples, storage, and computation. In addition, the representation and processing commitment made in the design of these methods have been shown to lead to inconsistent and arbitrary responses in practical domains characterized by substantial uncertainty, particularly when personalized decisions have to be made quickly in novel situations~\cite{kambhampati2024position,lu:acl24,vafa:icml25,xiong2024can}.

Many AI researchers are developing methods to address the known limitations of FMs in order to build better assistive AI agents. They are doing so by devoting additional resources (e.g., more and better curated data, more compute), building larger models, and developing better optimization methods. In a departure from this focus on data, compute, and optimization, we seek to explore the design of architectures for assistive AI agents by drawing on observations about natural and artificial systems, which can be traced back to the early pioneers of AI but are not being leveraged fully by modern AI methods~\cite{sridharan2025back}. This includes the key observation that the representation and processing commitments made in the design of an architecture (pre)determine the capabilities of the architecture~\cite{mccarthy:MI69}. Also, supporting a diverse set of representations and processing methods, and rapidly directing attention to the subset relevant to the tasks and domain at hand are essential capabilities for robust operation in machines and humans~\cite{broadbent:PR57,mccarthy:MI69,triesman:CP80}. In addition, humans use simple heuristics based on certain core capacities (e.g., to sum, order, imitate, recognize, forget, track) to rapidly acquire predictive models of behavior and make robust decisions \emph{in the wild}, i.e., under conditions of uncertainty and intractability~\cite{simon:PR56}. The use of such heuristics has also led to good performance in many practical application domains~\cite{gigerenzer:MMM16}. The early pioneers of AI mapped these observations to fundamental principles such as iterative refinement (i.e., compositionality), ecological rationality, interactive learning, and integrated reasoning and learning. They also advocated for these principles to be embedded as the building blocks of computational architectures developed for AI agents. Unlike work that attempts to insert or discover such principles in deep networks, our long-term objective is to explore a broad range of representation and processing commitments for introducing these principles in the design of the basic building blocks of architectures for assistive agents, with modern AI methods (e.g., FMs) being one of many different tools that these agents can choose from.

\begin{figure}[t]
    \centering
    \includegraphics[width=1\columnwidth]{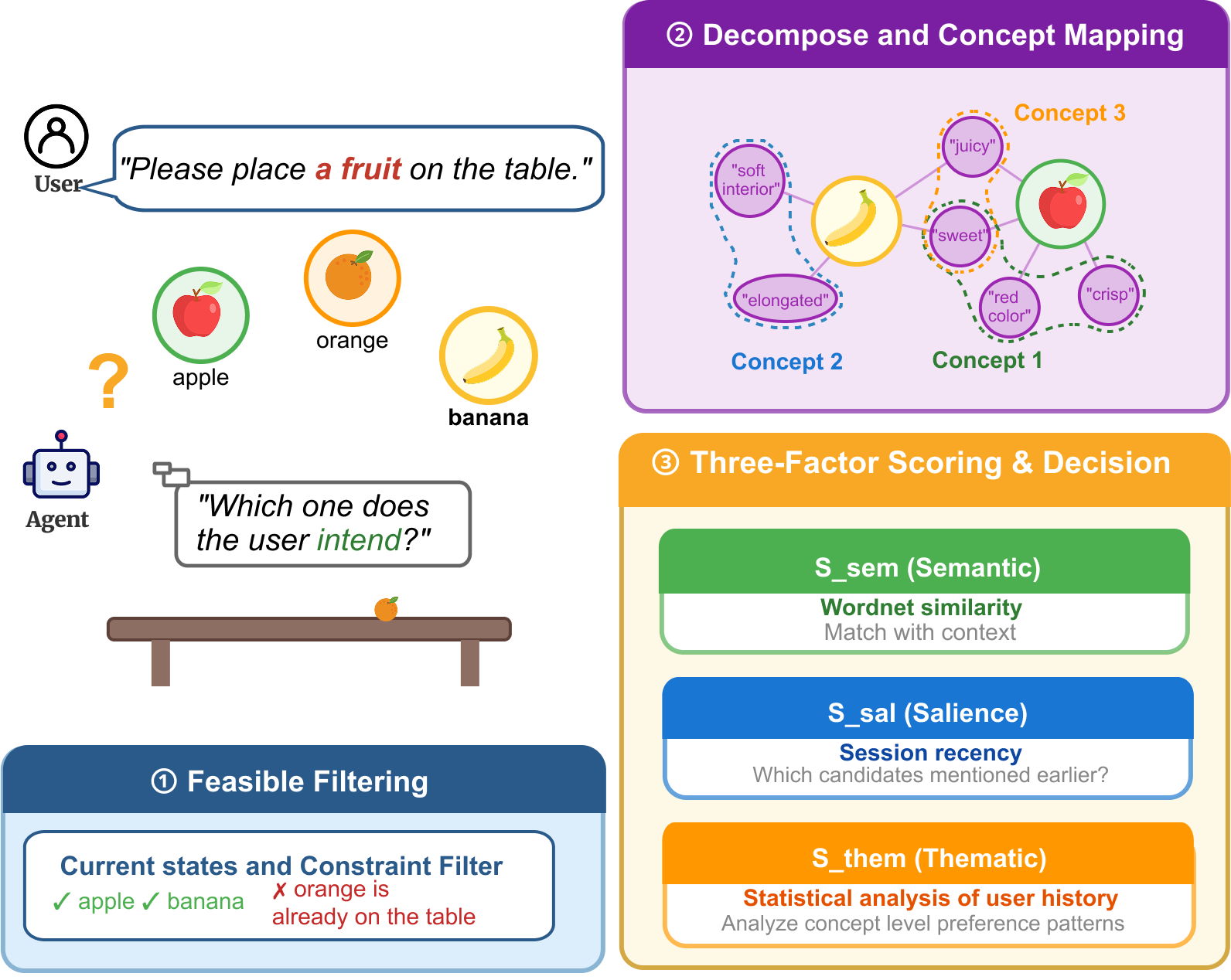}
    \vspace{-1.5em}
    \caption{Architecture overview and illustrative example of an AI agent receiving an ambiguous request. Our architecture enables the agent to reason with prior knowledge, learned compositional concepts, and contextual information modeled using three heuristic factors, to identify and score valid candidates, resolving ambiguous references or triggering clarification as needed. Attribute labels in the concept-mapping panel are shortened for display; the full attribute strings are \emph{fruit is juicy}, \emph{can be red colored}, and \emph{has a soft interior}.}
    \label{fig:motivating}
    \vspace{-1em}
\end{figure}

The representative use case considered in this paper is that of an AI agent that has to resolve the ambiguity in the objects being referred to by the human user, although disambiguation itself is not the objective of this work. Consider such an agent that has been asked by a human to ``place a fruit on the table'' in a kitchen containing apples, oranges, and bananas, as shown in Figure~\ref{fig:motivating}. Such ambiguous statements are common in human conversations, and we achieve disambiguation using prior and acquired knowledge of domain and the preferences and habits of the human(s) we are interacting with. It is challenging for AI agents based on deep networks to achieve such personalized disambiguation rapidly. Interactive clarification may help resolve this ambiguity, but answering many questions from the AI agent will be a cognitive burden for the human user. Some hybrid frameworks combining data-driven models and knowledge-based reasoning have also been explored in an attempt to address ambiguity in task specification~\cite{guan2023leveraging,lin2024clmasp}, but effective personalized interaction remains an open problem.

%%%%%%%%%%%%%%%%%%fig1%%%%%%%%%%%%%%%%%%%%%%%%%%%
%%%%%%%%%%%%%%%%%%fig1%%%%%%%%%%%%%%%%%%%%%%%%%%%

In the architecture described in this paper, we focus on embedding the principle of hierarchical compositionality and the use of simple heuristics, and thoroughly explore the interplay between reasoning and learning in the associated components. Based on the chosen use case, we limit the compositional hierarchy to representing and leveraging information about domain objects in the form of physical and functional attributes at different (linked) abstractions. Such a hierarchy has been shown to support fast and reliable learning and adaptation in new situations (see Section~\ref{sec:relwork}). For example, knowledge of a user's regular consumption of oranges in the afternoon can also be considered as a preference for fruits and healthy options for an afternoon snack. If the AI agent is then asked to fetch a snack but there are no oranges available, it can use prior knowledge to offer the user apples instead of chips. Future work could extend the  compositional representation to temporal events, representing (say) the observed movement of a dog in terms of changes in the parts of the dog and using this knowledge to generate plausible motion for a previously unseen four-legged animal. In the design and use of our compositional hierarchy, we embed simple heuristics based on some core capacities. We then explore the impact of learning and leveraging such a hierarchy of domain objects on disambiguating commands provided by a human interacting with an AI agent. Specifically, our architecture:
\begin{itemize}
    \item Represents entities as hierarchical compositions of atomic attributes drawn from human-validated semantic feature norms and compound \textit{concepts} automatically mined from these attributes, with personalized preferences of human users learned from observations of their behavior.
    % Represents entities as hierarchical compositions of automatically selected atomic and compound physical attributes learned from observations to capture personalized preferences.% and discover user-specific habit patterns at the concept level, enabling preference evidence to transfer across entities used in different tasks and situations.
    
    \item Computationally and heuristically models semantic compatibility, session salience, and user-specific thematic preference as complementary factors that capture contextual information about domain objects and the human users' interaction with them. 
    
    \item Performs non-monotonic logical reasoning with current knowledge of concept hierarchy, axioms governing domain dynamics, and computed model of contextual information to achieve disambiguation, requesting clarification from humans only when needed.
\end{itemize}
We use Answer Set Programming (ASP)~\cite{gelfond2014knowledge} as the non-monotonic logical reasoning paradigm, and use GPT-5.1~\cite{singh2025openai} as the LLM. We ground and evaluate our architecture's capabilities in the context of an assistive agent aiding humans with different user profiles in a realistic physics-based simulated environment. We use qualitative and quantitative results from ablation studies to demonstrate clear performance gains over state of the art LLM baselines. Furthermore, we highlight the important contributions made by different components of our architecture by demonstrating clear performance gains over baselines that use various subsets of our architecture's components.

The remainder of the paper is organized as follows. We first motivate the need for our architecture based on a review of related work (Section~\ref{sec:relwork}). We then describe the problem formulation and our architecture (Section~\ref{sec:problem-arch}). This is followed by a description of the experimental setup and results (Section~\ref{sec:exp-setup-results}), and a discussion of conclusions and future work (Section~\ref{sec:conclusion}).

%%%%%%%%%%%%%%%%%%%%%%%%%%%%%%%%%%%%%%%%%%%%%%%%%%%%%%%%%%%%%%%%%%%
%%%%%%%%%%%%%%%%%%%%%%%%%%%%%%%%%%%%%%%%%%%%%%%%%%%%%%%%%%%%%%%%%%%
\section{Related Work}
\label{sec:relwork}
We begin by reviewing work in command disambiguation, user modeling, and compositional representations in the context of AI agents assisting humans.

%%%%%%%%%%%%%%%%%%%%%%%%%%%%%%%%%%%%%%%%%%%%%%%%%%%%%%%%%%%%%%%%%%%
\subsection{Command Disambiguation and Personalization}
\label{sec:relwork-disambiguate}
% With the increasing use of FMs in Human-Robot Interaction (HRI), there has also been related work on addressing ambiguity~\cite{tellex2020robots,yu2018mattnet,deng2021transvg,huang2022language,ahn2022can,hemanthage2024divide}. This includes work on matching referring expressions to visual regions~\cite{yu2018mattnet,deng2021transvg,hemanthage2024divide} and prompting LLMs to classify command ambiguity and generate clarification questions~\cite{park2023clara}. However, these methods rely solely on information encoded in the FMs, visual features extracted from the scene, or dialogue history, without modeling individual user preferences. When multiple candidates satisfy the linguistic description, they lack sufficient signals to identify the intended referent. The fact that LLMs rely on statistical associations learned from large datasets also leads to poor performance when personalization is required~\cite{kambhampati2024position,xiong2024can}.

With the increasing use of FMs in Human-Robot Interaction (HRI), there has been work on addressing ambiguity in such models~\cite{tellex2020robots,yu2018mattnet,deng2021transvg,huang2022language,brohan:corl23,hemanthage2024divide}. This includes work on matching referring expressions to visual regions~\cite{yu2018mattnet,deng2021transvg,hemanthage2024divide}, prompting LLMs to classify command ambiguity and generate clarification questions~\cite{park2023clara}, and converting the cross-modal attention of a VLM into a spatial signal that detects referential ambiguity and triggers clarification~\cite{abrini2026clue}. However, these methods rely solely on information encoded in the FMs, visual features extracted from the scene, or dialogue history, without modeling individual user preferences. When multiple candidates satisfy the linguistic description, they lack sufficient user-specific signals to rapidly identify the intended referent.

Clarification-based approaches address ambiguity by posing questions to the user, ranging from enumerating individual candidates~\cite{shridhar2020ingress,hatori2018interactively} to selecting discriminative attributes or semantically informative questions for more targeted queries~\cite{thomason2019improving, jiang2024llms,dogan2025model}. We instead draw inspiration from a framework that leverages the interplay between knowledge-based reasoning and data-driven learning, reasoning with prior knowledge to eliminate unsuitable candidates by constructing clarification queries~\cite{mota2021answer}. However, even this framework relies on substantial human supervision, and it does not have the hierarchical compositional representation of knowledge that forms a key component of our architecture.

% A common limitation of much of the existing work based on modern AI methods is that the disambiguation method does not build on user-specific cues. In practice, however, different users often intend different referents for the same ambiguous command. Recent work has begun to address this by recording explicit user preferences through LLM conversation memory~\cite{abugurain2024integrating}, but such approaches require users to verbally state their preferences and cannot transfer evidence to unseen entities. Our architecture enables the AI agent to learn and reason with models of user-specific preferences that are linked to the learned compositional concepts, requesting clarification from the human user only when necessary.

A common limitation of much of the existing work based on modern AI methods is that the disambiguation method does not build on user-specific cues. In practice, however, the same ambiguous command may refer to different entities for different users and in different contexts. The reliance of LLMs on statistical associations learned from large datasets leads to poor performance when such personalization needs to be achieved quickly~\cite{kambhampati2024position,xiong2024can}. Even recent benchmarks assume no access to user history, either confining dialogue and questions to the current scene within a single task episode~\cite{gao2022dialfred,padmakumar2022teach} or deliberately imposing a zero-context setting in which every preference-type ambiguity must be resolved by asking the user~\cite{ivanova2025ambik}. The approaches that exploit user-specific evidence either reuse the past referring expressions recorded for each object~\cite{roy2019leveraging} or retain explicitly stated preferences in an LLM memory~\cite{abugurain2024integrating}. These strategies do improve disambiguation for entities with recorded history, but do not transfer the accumulated evidence across entities. Our architecture enables the AI agent to reason and learn with models of user-specific preferences that are linked to the learned compositional concepts, requesting clarification from the human user only when necessary.

More broadly, personalization to individual users continues to be an open problem for robots and agents controlled by FMs~\cite{wu2025aligning}. Existing approaches learn behavioral patterns from interaction history through explicit feedback signals~\cite{christiano2017deep,wirth2017survey} or implicit signals in the form of command history~\cite{fischer2001user,fu2025combining}. 
%new added 10/06
Beyond command-level interactions, research has shown that stable per-user regularities can also be recovered from sequences of everyday behavior. The smart-home studies cluster recognized daily-activity sequences and recovered recurring routines specific to each user~\cite{sepesy2021discovering}. However, these routines were primarily mined for activity monitoring and anomaly detection rather than for interpreting subsequent user commands. Other work has focused on anticipating future behavior from egocentric video, e.g., predicting the next object a person will interact with~\cite{furnari2017next} or recommending which assistive action should be invoked~\cite{abreu2024parse}. These studies show that user histories contain predictive regularities, but they do not use them to resolve entity-level ambiguity in user commands.
In addition, research has explored transferring user preferences across assembly tasks by learning from demonstrations in a simpler canonical task and adapting the acquired knowledge to more complex tasks~\cite{nemlekar2023transfer}. In practical domains, high-level tasks are often linked to the specific target entities to be manipulated. However, there is a scarcity of data needed to acquire preferences at the (low) level of primitive actions and specific entities. Our architecture addresses this limitation by acquiring a hierarchical compositional representation of objects, enabling transfer of knowledge across related entities from limited interaction history.

%%%%%%%%%%%%%%%%%%%%%%%%%%%%%%%%%%%%%%%%%%%%%%%%%%%%%%%%%%%%%%%%%%%
\subsection{Compositional Hierarchical Representations}
\label{sec:relwork-compositional}
Hierarchical compositional representations have been studied for many decades in the context of human cognition~\cite{fodor:book75,knoblich:PS01,piantadosi:PR16} and probabilistic sequential decision making~\cite{dietterich:icml98}; they have also received an information-theoretic grounding~\cite{elmoznino:icml25}, and used in different domains. In visual recognition, low-level features were recursively composed into shared parts across object categories~\cite{fidler2007towards}, and shared semantic properties enabled zero-shot transfer to unseen categories~\cite{lampert2013attribute}. In robotics, encoded and learned ontologies (and knowledge graphs) have been used for planning~\cite{beetz:icra18}, kinematic primitives have been hierarchically composed into transferable motor representations~\cite{vzabkar2016motor} and motion primitives are composed through graph structures for novel tasks~\cite{tian2024gsc}. This principle has also been applied to referring expressions in human-robot interaction~\cite{gao2023compositional} and preference modeling~\cite{go2023compositional}. Our work is directed at understanding if and how a compositional hierarchy can be acquired and used in architectures governing AI agents. In this paper, we do so in the context of disambiguating user commands. 

%%%%%%%%%%%%%%%%%%%New added related work for deep models 

Researchers focusing on modern AI methods based on deep networks have also explored compositionality. Some studies argue that complex target functions contain reusable lower-dimensional components that deep networks can exploit~\cite{danhofer2025position}, while others show that deep networks can learn composed functions without assuming such an explicit decomposition in advance~\cite{jacot2025how}. Other studies have examined when such structure is actually learned and exploited by deep models. For example, prior work shows that deep networks can generalize compositionally when the data contains learnable structure, the model can align with this structure, and the training distribution covers the relevant combinations~\cite{Cagnetta2024HowDeep,boopathy2024breaking,redhardt2026scaling}. However, these deep networks and LLMs may still rely on seen combinations or statistical shortcuts rather than systematically combining learned concepts~\cite{lippl2025when}. They can often answer the individual sub-questions of a task but fail when the task requires composing them in a specific manner, and this compositionality gap does not necessarily close as models are provided additional data~\cite{Press2023Ofir}. Even when the final output is correct, mechanistic studies show that the model's internal computation may not capture the intermediate results~\cite{khandelwal2026how}.

A related line of research attempts to make the intermediate structure of deep network models more explicit. For example, concept-based models introduce a concept layer so that the deep network first predicts human-interpretable concepts and then uses them for classification or prediction~\cite{pmlr-v119-koh20a}. Neurosymbolic methods combine neural perception with logical or probabilistic inference to enable use of concepts or symbolic variables in downstream reasoning~\cite{chen2025neural}. These methods often rely on predefined concepts and concept-level annotations, shifting much of the modeling burden to humans and limiting adaptation to fine-grained, user-specific preferences in changing contexts. Also, they are developed primarily for classification or prediction in (near)static conditions, rather than for rapid adaptation to specific users and dynamic environments. Our approach lies between fully implicit neural representations and explicit symbolic representations. The basic set of attributes is chosen from an open-source dataset of human-validated features, but the concept-level structures above it are learned from attribute co-occurrence and user interaction history, allowing our architecture to capture subtle preferences that are difficult to specify in advance. These structures can be audited and revised, and be combined with action constraints, context, and history for resolving ambiguous instructions.

%%%%%%%%%%%%%%%%%%%%%%%%%%%%%%%%%%%%%%%%%%%%%%%%%%%%%%%%%%%%%%%%%%%
\subsection{Decision Heuristics}
\label{sec:relwork-heuristics}
There is a rich history on the design and use of heuristics in different disciplines; they are often viewed as shortcuts used to find solutions quickly or as approximations of optimal solutions. The early pioneers of AI and modern mathematics, on the other hand, drew inspiration from the use of heuristics by humans to propose strategies for solving intractable problems~\cite{katsikopoulos:JMM26}. One example of a related computational theory is Herb Simon's work on \textit{Bounded Rationality} and heuristics~\cite{simon:PR56}.%this also serves as our interpretation of heuristics in this paper. 

Over time, bounded rationality has been incorrectly equated with maximizing expected utility (e.g., finance, computing)~\cite{russell:aibook03}. These methods actually focus on optimal search in the presence of \textit{risk}, i.e., a search over a set of known states and outcomes. This includes methods for logical reasoning and probabilistic reasoning, and the modern AI methods based on FMs and deep networks. Another interpretation of bounded rationality is to be found in the \textit{heuristics and biases} program in Psychology, where the use of heuristics by humans came to be viewed as leading to biases in their decisions~\cite{tversky:SCIENCE74}, leading to well-known dual-process theories~\cite{kahneman:book11,deneys:book18} that have been used extensively in AI~\cite{gronchi:FC24}. More recent work has cast heuristics as priors to more complex optimization methods, or as Bayesian inference under limit of infinitely strong priors, but it only explores some basic heuristics and simplistic problem domains under strong assumptions about knowledge of states and outcomes~\cite{parpart:CP18}. Research in Psychology and other disciplines has shown that these interpretations do not match the original definition of bounded rationality, have limited support in human decision making, do not account for "less is more" effects under uncertainty, and do not provide insights into the process of making decisions~\cite{gigerenzer:MMM16,katsikopoulos:JMM26}. 

Our work builds on the principle of \textit{ecological rationality}, which is based on the original (i.e., Herb Simon's) definition of bounded rationality and heuristics~\cite{gigerenzer:bookchap20}. The focus here is on decision making under \textit{open world uncertainty} or "in the wild", i.e., when the space of possible states and outcomes is not known in advance. In such circumstances, the optimal solution is ill-defined and probabilistic estimates may cease to be meaningful. The behavior of a human or an AI system is characterized as a joint function of the internal cognitive processes and the environment, and \textit{adaptive satisficing} is used as the guiding principle to design \textit{simple heuristics} such as tallying, lexicographic search, and fast and frugal (FF) trees in order to rapidly learn (and revise) predictive models and make rational decisions. Instead of hacks or biases, heuristics are viewed as leveraging core (human) capacities such as counting, ordering, tracking, and collaborating, in order to ignore part of the information and make decisions more quickly, frugally, and accurately than more complex methods~\cite{gigerenzer:MMM16}. 

The difference between modern AI methods and simple heuristics can be understood in terms of the \textit{bias-variance decomposition} of the expected value of the squared prediction error~\cite{geman:NC92}. Modern AI methods are based on the belief that a large number of free parameters are necessary to generalize to different situations. Having many tunable parameters can lead to overfitting, with a low value of the bias component of the error and a high value of the variance component of the error resulting in high (overall) error, particularly in novel situations. Simple heuristics, on the other hand, are based on limited free parameters. They may lead to higher value of the bias component but still have lower overall error compared with modern AI methods because of the much lower variance component of the error; even the bias component is small in many practical domains in which properties such as dominance and non-compensatoriness hold~\cite{simsek:bookchap20}. Unlike modern AI methods that are largely \textit{prescriptive}, i.e., they describe what \emph{should} be done in specific situations to achieve specific outcomes, decision heuristics are both prescriptive and \textit{descriptive}, i.e., they also capture what people or agents do to achieve specific outcomes. This descriptive capability supports the automatic generation of process-level descriptions as explanations of decisions. In the long-term, our work seeks to develop an adaptive toolbox of classes of such heuristics, FM, and other methods. Given any particular problem, we can then match the characteristics of the problem with those of classes of tools, along with an algorithmic approach and out-of-distribution testing, to identify the \textit{tool} suitable for the problem.  

Although decision heuristics have provided good performance on prediction problems in application domains such as finance,
healthcare, and law~\cite{brighton:ER12,durbach:DSS20,gigerenzer:MMM16,katsikopoulos:IJF21}, there is limited use of these methods in architectures for robots or AI agents,
except in some related work in the cognitive systems community~\cite{langley:acs22} and prior work in our group~\cite{dodampegama:FAI26}. This lack of attention to decision heuristics is potentially because their simplicity makes researchers doubt their suitability for addressing complex practical problems. In this work, we use heuristic methods to aid in the design and use of our compositional hierarchy.

% %%%%%%%%%%%%%%%%%%%%%%%%framework%%%%%%%%%%%%%%%%%%%%%%%%%%%%%%%%%%%%%%%%%%%%%%%%%%%%%%%%%%%%%%%
\begin{figure}[t]
    \centering
    \includegraphics[width=\textwidth]{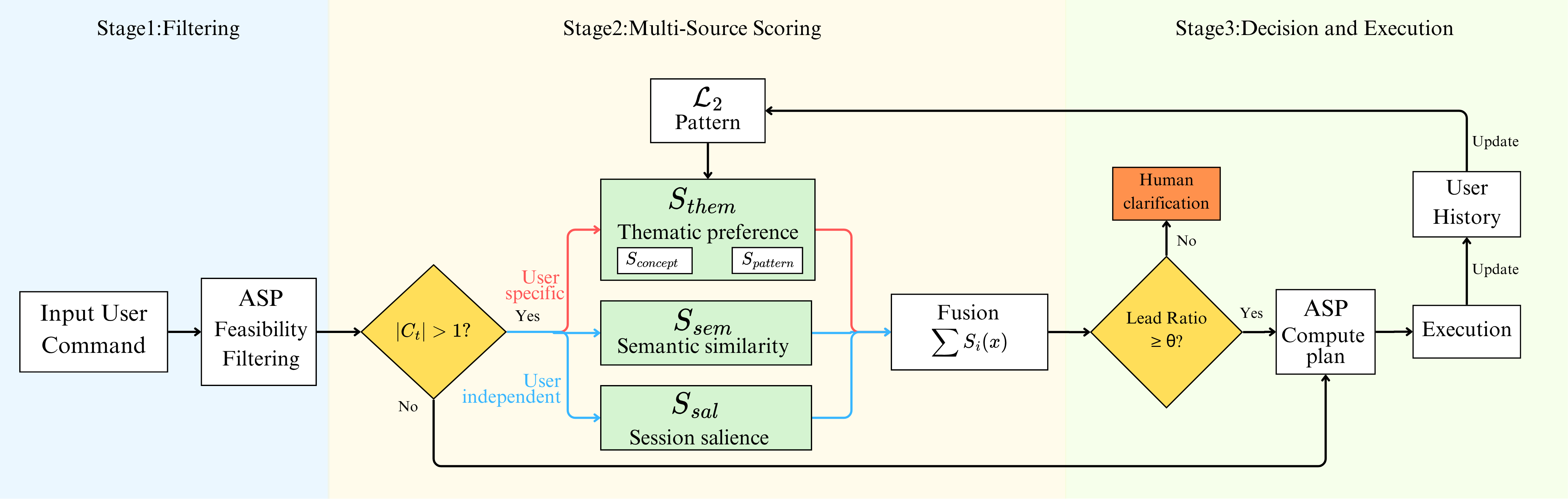}
    \caption{Three-stage disambiguation framework. \textbf{Stage~1:} The agent parses the input user command and applies ASP-based feasibility filtering using domain knowledge ($\Sigma$, $\Pi$) to obtain a candidate set $\mathcal{C}_t$. \textbf{Stage~2:} If $|\mathcal{C}_t|>1$, candidates are scored by fusing semantic similarity ($S_{sem}$), session salience ($S_{sal}$), and user-specific thematic preference ($S_{them}$). \textbf{Stage~3:} The agent selects the top candidate if the lead ratio exceeds threshold $\theta$, or invokes human clarification otherwise. Each resolved interaction updates the user history, enabling continuous refinement of learned user preferences.}
    
    \label{fig:framework}
\end{figure}
%%%%%%%%%%%%%%%%%%%%%%%%%%%%%%%%%%%%%%%%%%%%%%%%%%%%%%%%%%%%%%%%%%%%%%%%%%%%%%%%%%%%%%%%

%%%%%%%%%%%%%%%%%%%%%%%%%%%%%%%%%%%%%%%%%%%%%%%%%%%%%%%%%%%%%%%%%%%%%%%%%%%%%%%%%%%%%%%%%
\section{Problem Formulation and Architecture}
\label{sec:problem-arch}
Consider an embodied AI agent assisting a human in a home environment to perform tasks $\{\tau_1, \tau_2, \dots, \tau_n\}$ assigned sequentially.  A contiguous subsequence of thematically related tasks forms a 
\emph{session} (e.g., a sequence of commands constituting a routine for preparing for afternoon tea). Session boundaries define the scope of short-term contextual signals, e.g., salience evidence accumulates within a session and resets at each boundary. Each task $\tau_i$ is executed as a sequence of fine-grained actions such as \emph{pickup}, \emph{bring}, and \emph{open}. Disambiguation is required when: (a) the command contains an under-specified reference, such as a category term (e.g., ``the drink'') or a pronoun (e.g., ``it''), that is compatible with multiple entities in the current state; or (b) the command is incomplete (e.g., "please fetch the ...") because the human input was not processed properly.

Figure~\ref{fig:framework} outlines the flow of information and control in our architecture, organized in the form of three stages split over two phases. In the first stage, the agent first parses any given command to construct a candidate set $\mathcal{C}$ of feasible interpretations of the input command. This involves ASP-based non-monotonic logical reasoning (Section~\ref{sec:problem-kr}) with any prior knowledge (if available) of domain state and constraints to filter infeasible candidates. If there are multiple feasible candidates, they are then ranked in the second stage using three heuristically-modeled complementary signals (Section~\ref{sec:problem-disambig}): (i) user-independent semantic compatibility derived from WordNet~\cite{miller1995wordnet} and recent execution outcomes; (ii) session-level salience based on recent references within the current session; and (iii) user-specific thematic preference estimated from interaction history based on a hierarchical compositional representation (Section~\ref{sec:problem-compositional}). In the third stage, if there is no clear candidate identified to resolve the ambiguity, the agent solicits clarification from a human. Once the agent identifies a specific interpretation of the ambiguous command, ASP-based reasoning is used to compute and execute a task plan that executes the user command. This trace of interaction and observations is also added to the user-specific history, enabling subsequent revision of the learned user preferences. We now discuss the individual components of the architecture, along with the interplay between them, starting with the knowledge representation and reasoning component.

%It must then either select the user-intended referent $g_t \in \mathcal{C}_t$ or request clarification when the available evidence is insufficient. To address the inherent sparsity of interaction evidence at the entity level, we represent entities through a hierarchical composition of shared concepts. This enables preference evidence to be collected and transferred across entities with physical or functional similarity.

%The highest-ranked candidate is selected only when its lead exceeds a fixed confidence threshold; otherwise, the agent requests clarification.
%%%%%%%%%%%%%%%%%new addded as a loop
% Once an entity is resolved, whether through confident selection or after human clarification, the agent invokes ASP to compute and execute the task plan. The resulting interaction record is then appended to the user history, enabling continuous refinement of the learned user preferences.
%%%%%%%%%%%%%%%%%

%%%%%%%%%%%%%%%%%%%%%%%%%%%%%%%%%%%%%%%%%%%%%%%%%%%%%%%%%%%%%%%%%%%%%%%%%%%%%%%
\subsection{Knowledge Representation and Reasoning with ASP}
\label{sec:problem-kr}
In our architecture, any domain's transition diagram is represented using action language $\mathcal{AL}_d$~\cite{gelfond2013system}. Action languages are formal models of parts of natural language for describing prior knowledge about a domain's transition diagrams. The domain representation comprises a system description $\mathcal{D}$, a collection of statements of $\mathcal{AL}_d$, and a history $\mathcal{H}$. $\mathcal{D}$ has a sorted signature $\Sigma$ with basic sorts, and the domain attributes (statics and fluents) and actions are described in terms of these basic sorts.

Our domain includes basic sorts such as $room$, $container$, $furniture$, $food$, $human$, and $step$ (for temporal reasoning) that are arranged hierarchically, e.g., $fruit$ is a sub-sort of $food$ that is a sub-sort of $item$. Statics are domain attributes whose values cannot change, e.g., $next\_to(kitchen, study)$, and fluents are attributes whose values can change. Fluents can be \emph{inertial}, which obey inertia laws and are changed by actions, e.g., $loc(item, room)$ and $in\_hand(agent, item)$; and \emph{defined}, which do not obey inertia laws and are not directly changed by the agent's actions, e.g., $otherloc(human, room)$ is the human's location. Actions include the agent's actions, e.g., $move(agent, room)$, $pickup(agent, item)$, and $switchon(agent, appliance)$.

For any given domain, prior knowledge of the domain's dynamics is described based on the corresponding $\Sigma$ using three types of axioms:
\begin{subequations}
\label{eqn:axioms}
\vspace{-0.5em}
\begin{align}
    move(A, R) &~\mathbf{ causes }~ loc(A, R) \\
    \neg\, at(A, R_1) &~\mathbf{ if }~ at(A, R_2), R_1 \neq R_2\\
    \mathbf{impossible}~ give(A, O, U)~&\mathbf{if}~ loc(A, R_1), ~otherloc(U, R_2), R_1\neq R_2
\end{align}
\end{subequations}
where Statement~\ref{eqn:axioms}(a), a \textit{causal law}, implies that when an agent ($A$) moves to a room $R$, its location becomes $R$; Statement~\ref{eqn:axioms}(b), a \textit{state constraint}, implies that an agent ($A$) cannot be in two places ($R_1, R_2$) at the same time; and Statement~\ref{eqn:axioms}(c), an \textit{executability condition}, prevents the agent ($A$) from trying to give an object ($O$) to a human user ($U$) who is not in the same room as the agent. In addition to such a system description $\mathcal{D}$, the history $\mathcal{H}$ is a record of statements of the form $obs(fluent, boolean, step)$, which represent observations received at specific time steps, and of the form $hpd(action, step)$, which represent actions executed at specific time steps.

In our architecture, ASP-based non-monotonic logical reasoning is used for inference, planning and diagnosis. To perform such reasoning with knowledge, an ASP program $\Pi(\mathcal{D}, \mathcal{H})$ is automatically constructed (using a Python script) to include statements from $\mathcal{D}$ and $\mathcal{H}$, inertia axioms, reality check axioms, closed world assumptions for defined fluents and actions, and helper relations to reason over time steps, e.g., $holds(fluent, step)$ and $occurs(action, step)$ imply (respectively) that a fluent is true and that an action is part of a plan at a particular time step. For planning and diagnosis (in Stage 3),  $\Pi(\mathcal{D}, \mathcal{H})$ includes helper axioms to define goals and guide the search for plans or explanations (diagnosis). After executing each action in the plan, the corresponding $hpd(action, step)$ and any resulting observations of the form $obs(fluent, boolean, step)$ are appended to $\mathcal{H}$ for subsequent reasoning. Similarly, for inference and filtering of infeasible candidate interpretations of the user command (in Stage 1), the agent queries $\Pi(\mathcal{D}, \mathcal{H})$ to mentally simulate the execution of the command and compare the fluent values between current state and the resultant state. Any entity $o \in \mathcal{O}$ that appears as an argument of a fluent literal whose value has changed between the two states is identified and collected in a set $O^{\Delta}_t$ at time $t$. This set serves as contextual input for subsequent disambiguation (as described in Sections~\ref{sec:problem-compositional}--\ref{sec:problem-disambig}).

ASP is based on stable model semantics, and encodes \emph{default negation} and \emph{epistemic disjunction}; unlike classical negation ``$\lnot p$'', which states that \emph{p is believed to be false}, default negation ``$not~p$'' only implies that \emph{p is not believed to be true}. Each literal is true, false, or unknown, and the agent only believes that which it is forced to believe. ASP supports non-monotonic reasoning, i.e., the ability to revise previously held conclusions, which is essential for agents operating with incomplete knowledge and noisy observations. All reasoning tasks are reduced to computing \textit{answer sets} of $\Pi$, and we use the SPARC system~\cite{balai2013towards} to compute these answer sets. For example programs corresponding to our paper, please see~\cite{code-results}.

\medskip
\noindent
\textbf{Need for compositional hierarchy and heuristics.}
Although ASP filtering removes candidates violating known domain constraints, multiple feasible interpretations of the command often remain. These interpretations are based on the different objects (i.e., entities) that the ambiguous (or missing) part of the command may be referring to. It is difficult to identify and rank these feasible candidates, particularly if we are to consider domain-specific constraints and user preferences, because knowledge of the domain and the history ($\mathcal{H}$) of interactions involving any specific user may be limited. Even if additional data becomes available, it is difficult to determine which parts of this data need to be combined (and how) to provide the desired personalized response. Simple heuristics are well-suited for making decisions in such circumstances. In our architecture, the design and use of heuristics is driven by two simple rules:
\begin{enumerate}
    \item Describe any (new) object in terms of attributes commonly used (by humans) to describe previously known objects in the domain, and in terms of abstract concepts that capture attribute co-occurrences.

    \item Identify the most likely object being referenced by matching attributes of candidate objects with those of known objects in recent interactions, including those involving this specific user.
\end{enumerate}
%In an attempt to address the scarcity of data needed to resolve ambiguities in complex domains, our architecture 
While it is often claimed that FMs and deep networks capture statistics of object co-occurrence, it is difficult for such models to provide the desired personalization under uncertainty. Our work instead maps the first rule above to a simple compositional hierarchy of attributes and concepts that can be acquired and revised rapidly (Section~\ref{sec:problem-compositional}). Based on this representation, our architecture then maps the second rule to heuristic measures of \textit{semantic similarity}, \textit{salience}, and user-specific \textit{thematic preferences} (Section~\ref{sec:problem-disambig}). These measures are based on well-known simple heuristics such as the \textit{recognition heuristic}, which advocates for choosing that which we know already over new entities~\cite{goldstein:PR02}, and the \textit{weighted tallying heuristic} that combines a simple set of cues by adding evidence from each cue weighted by (learned) cue importance~\cite{gigerenzer:MMM16}. We hypothesize and experimentally demonstrate that such an approach provides much better performance than the large, monolithic modern AI models while requiring orders of magnitude fewer resources.

\begin{figure}[t]
    \centering
    \includegraphics[width=\textwidth]{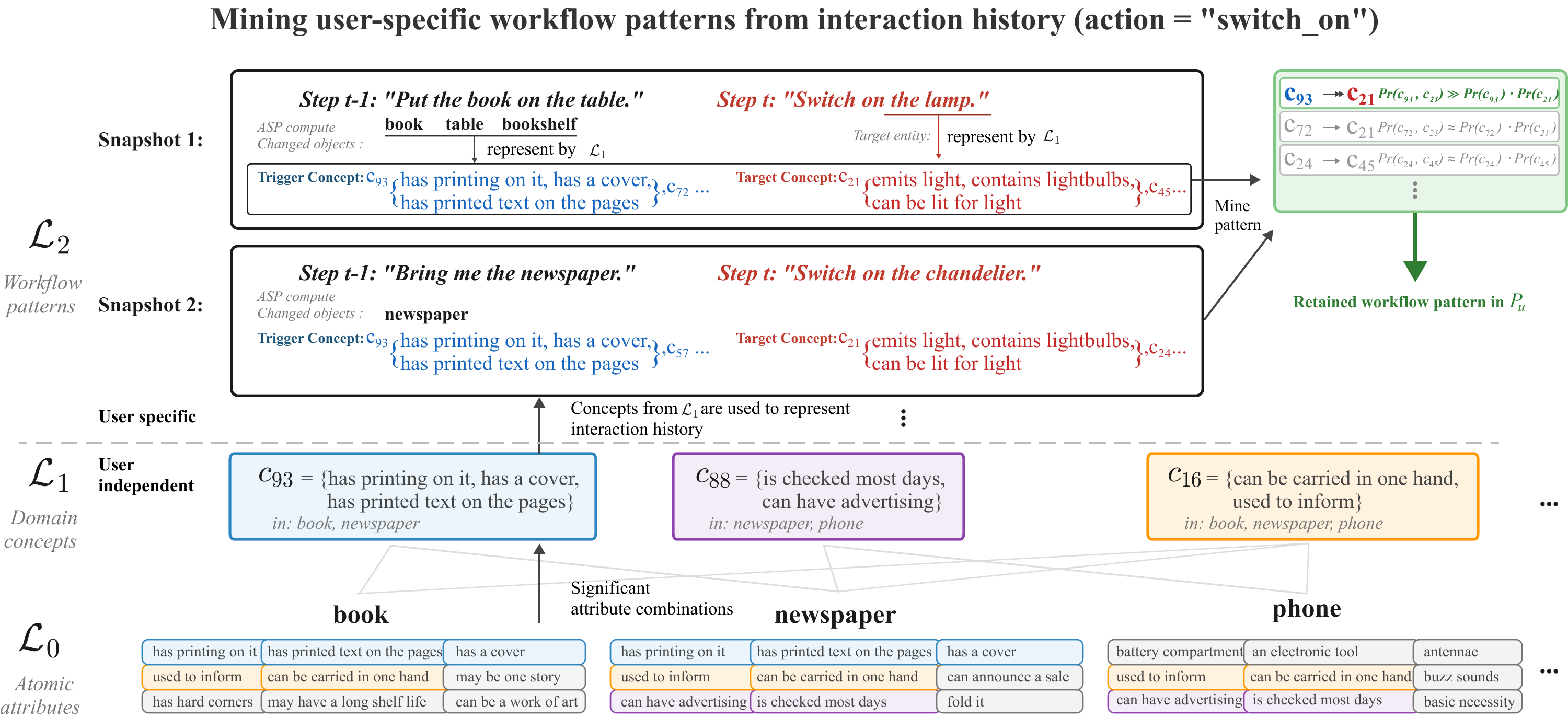}
    \caption{Overview of the three-layered hierarchical and compositional representation. $\mathcal{L}_0$ and $\mathcal{L}_1$ are user-independent and domain-specific atomic attributes and concepts (respectively), with attributes drawn from human-validated semantic feature norms and concepts mined from significant attribute combinations; $\mathcal{L}_2$ captures user-specific workflow patterns from interaction snapshots filtered by the action predicate of the current command. Figure illustrates two example snapshots for action \textit{switch on}, with the retained pattern $c_{93} \to c_{21}$ indicating that this user tends to interact with entities that emit light and contain lightbulbs after interacting with printed reading material such as the book and the newspaper.}
    \label{fig:compositional_hierarchy}
\end{figure}

%%%%%%%%%%%%%%%%%%%%%%%%%%%%%%%%%%%%%%%%%%%%%%%%%%%%%%%%%%%%%%%%%%%%%%%%%%%%%%%%%%%compositional_hierarchy%%%%%%%%%%%%%%%%%%%%%%%%%%%%%%%%%%%%%%%%%%%%%%%%%%%%%%%%%%%%%%%%%%%%%%%%%%%%%%%%%%%%%%%%%%%%%%%%%%%%%%%%%%%%%%%%%%%%%%%%%%%%%%%%%%%%%%%%%%%%%%%%%%%%%%%%%%%%%%%%%%%%%
\subsection{Compositional Hierarchy of Attributes and Concepts}
\label{sec:problem-compositional}
We first describe the three-layered hierarchical and compositional representation of objects in terms of their physical (appearance, functional) attributes. This representation forms the foundation and enables the approach for ranking the available candidates. The three layers of the representation correspond to:
\begin{itemize}
    % \item $\mathcal{L}_0$: domain objects (entities) are decomposed into fine-grained, shared, atomic attributes that capture the structure, shape, and physical attributes using a pretrained LLM; 
    \item $\mathcal{L}_0$: domain objects (entities) are automatically represented in terms of fine-grained, shared, atomic attributes drawn from NOVA, an open-source semantic feature norm dataset~\cite{suresh2025nova};

    \vspace{0.25em}
    \item $\mathcal{L}_1$: statistically significant attribute combinations are identified automatically to form a  domain-level concept library $\mathcal{K}$, with entities then being represented in terms of elements of this concept library; and 

    \vspace{0.25em}
    \item $\mathcal{L}_2$: entities are further represented in terms of user-specific workflow patterns and concept co-occurrences from records in $\mathcal{H}$ that correspond to actions (i.e., $hpd$ literals) and match the current command.
\end{itemize}
Note that $\mathcal{L}_0$ and $\mathcal{L}_1$ are user-independent and domain-specific; they are constructed before deployment. Also, these two levels can be revised quickly over time but we keep them fixed for the work described in this paper. $\mathcal{L}_2$, on the other hand, is user-dependent; it is constructed by gathering evidence dynamically, with each user maintaining a separate pattern set for each action (please see Figure~\ref{fig:compositional_hierarchy} for examples). We describe each layer (i.e., level) of the compositional hierarchy below.

%%%%%%%%%%%
\noindent
\paragraph{$\mathbf{\mathcal{L}_0}$\textbf{: Atomic Attributes.}}
% Let $\mathcal{O}$ be the set of all entities in the domain. For each entity $o \in \mathcal{O}$, we decompose it into a set of atomic attributes $A(o) = \{a_1, a_2, \dots, a_m\}$ across three dimensions: 
% \begin{enumerate}
%     \item \emph{structure}-based attributes such as layered, bound, solid, stacked, and sealed;
%     \item \emph{shape}-based attributes such as cylinder, cuboid, flat, compact, and thick; and 
%     \item \emph{physical} attributes such as plastic, metallic, lightweight, smooth, and rigid.
% \end{enumerate}
% In our architecture, a pre-trained LLM is used as an annotation tool for identifying these atomic attributes.
% %New added
% Specifically, for each entity, the LLM generates at least three attributes per dimension based on the entity label. Attributes that are identified for only one entity are excluded, as they cannot contribute to cross-entity transfer of knowledge. 
% %New added
% The process is repeated if necessary until each entity has a sufficient number of atomic attributes across each dimension. The union $\mathcal{A} = \bigcup_{o \in \mathcal{O}} A(o)$ then defines the full vocabulary of atomic attributes for the domain. At the end of this process, the attributes in $\mathcal{L}_0$ are fine-grained and universal, e.g., \emph{cylinder} appears across mugs, bottles, thermoses, and cans. However, these shared attributes lack the selectivity required to isolate specific functional preferences, motivating the discovery of higher-level (composite) attributes.
Let $\mathcal{O}$ be the set of all entities in the domain. Each entity $o \in \mathcal{O}$is represented as a set of atomic attributes $A(o) = \{a_1, a_2, \dots, a_m\}$, capturing the properties that people regularly judge the entity to possess. These attributes are drawn from \emph{semantic feature norms} in cognitive science, i.e., features collected by asking human participants to describe a concept's characteristic properties~\cite{rosch1975,mcrae2005}. We use NOVA~\cite{suresh2025nova}, a publicly-available dataset of human-validated feature norms covering $786$ everyday concepts. These attributes describe objects as humans perceive and name them, e.g., \emph{emits\_light} for a lamp and \emph{has\_printed\_text\_on\_the\_pages} for a book. Evidence can thus transfer across entities that share attributes that would be considered (by a human) to indicate similarity.

Attributes that are identified for only one entity are excluded, as they cannot contribute to cross-entity transfer of knowledge. The union $\mathcal{A} = \bigcup_{o \in \mathcal{O}} A(o)$ then defines the full vocabulary of atomic attributes for the domain. At the end of this process, the attributes in $\mathcal{L}_0$ are fine-grained and shared, e.g., \emph{emits\_light}, \emph{contains\_lightbulbs}, and \emph{can\_be\_lit\_for\_light} appear across the lamp and the chandelier, while \emph{has\_printed\_text\_on\_the\_pages}, \emph{has\_printing\_on\_it}, and \emph{can\_be\_carried\_in\_one\_hand} appear across the book and the newspaper. However, these shared attributes lack the selectivity required to isolate specific functional preferences, motivating the
discovery of higher-level (composite) attributes.

%%%%%%%%%%%%%%%%%%%%%%%%%%%%%%%%%%%%
\noindent
\paragraph{$\mathbf{\mathcal{L}_1}$\textbf{: Domain-level Attributes.}}
The next level in our hierarchy of attributes ($\mathcal{L}_1$) is a library $\mathcal{K}$ of statistically significant composite attributes, which we refer to as \textit{concepts}. 
% Starting from attribute pairs $\{a_i, a_j\} \subseteq A(o)$, we retain a composition only if it satisfies two contrasting terms:
% (i) \textbf{Significance}: the co-occurrence probability is significantly higher than predicted by random chance, i.e., $\Pr(a_i, a_j) \gg \Pr(a_i) \cdot \Pr(a_j)$, reflecting a structural archetype rather than random coincidence; and 
% (ii) \textbf{Frequency}: the combination is shared by a minimum number of entities ($|\{o \in \mathcal{O} \mid \{a_i, a_j\} \subseteq A(o)\}| > \theta$) to ensure evidence transferability.
% These concepts capture functional groupings that often lack standard category names (e.g., $\{\textit{cylinder, hollow, handle}\}$ clusters mugs and measuring cups but excludes thermoses). Each entity $o$ is then mapped to its concept set $\Gamma(o) = \{k \in \mathcal{K} \mid k \subseteq A(o)\}$. 
%%%%%%new added%%%%%%
In our architecture, the process for identifying these concepts starts by considering pairs of attributes $\{a_i, a_j\} \subseteq A(o)$, where $a_i$ and $a_j$ may belong to the same or different dimensions, and retains pairs (i.e., compositions) that satisfy two heuristically-modeled criteria:
% (i) \textbf{Significance}: the co-occurrence probability is significantly higher than predicted by random chance, i.e., $\Pr(a_i, a_j) \gg \Pr(a_i) \cdot \Pr(a_j)$; and 
\begin{enumerate}
    \item \textit{Significance}: this criterion seeks to identify attributes whose joint occurrence is more significant than the occurrence of the individual attributes. To do so, we first focus on attribute pairs and adapt the \emph{lift} measure defined in the data mining literature~\cite{brin1997dynamic}: 
    \begin{align}
        \label{eqn:lift}
        \mathrm{lift}(a_i, a_j) = \Pr(a_i, a_j) \,/\, (\Pr(a_i) \cdot \Pr(a_j)) \geq \lambda_1   
    \end{align}
    where $\lambda_1$ is an experimentally-determined satisficing threshold above which the joint occurrence of the attributes being examined is considered to be significant. 

    \vspace{0.25em}
    \item \textit{Frequency}: this criterion seeks to identify attribute pairs that have support from a minimum number of entities in the domain. It is computed in the form of a simple count of objects (with the attribute pair under consideration) exceeding a threshold:
    \begin{align}
        |\{o \in \mathcal{O} \mid \{a_i, a_j\} \subseteq A(o)\}| > \theta    
    \end{align}
    where $\theta$ is an experimentally-determined satisficing threshold. Together, these two criteria help identify attribute pairs shared across entities.
\end{enumerate}
Next, the attribute pairs that pass both criteria are incrementally extended by considering additional attributes, with each such combination re-evaluated using the same criteria, conditioned on the already-confirmed subset. For example, $\mathrm{lift}(a_k \mid \{a_i, a_j\}) = \Pr(a_i, a_j, a_k) \,/\, (\Pr(a_i, a_j) \cdot \Pr(a_k)) \geq \lambda_1$. This process continues until no further attribute passes both criteria or a predefined maximum size is reached for the concept. In our experiments, we set this upper bound on the size of each concept to five to limit computational cost while also identifying useful concepts. This yields concepts of varying sizes (e.g., from pairs such as $\{\textit{has\_printing\_on\_it, has\_printed\_text\_on\_the\_pages}\}$ and triples such as $\{\textit{has\_printing\_on\_it, has\_printed\_text\_on\_the\_pages, has\_a\_cover}\}$), each representing a more specific archetype to be considered for subsequent analysis.

In practice, a single entity is often represented by multiple concepts in $\mathcal{K}$, providing multiple pathways for evidence to transfer across structurally related entities. Conversely, attributes that do not participate in any retained combination are not covered by $\mathcal{K}$. %In the limiting case where $\Gamma(o) = \emptyset$, the thematic signal $S_{them}$ contributes no evidence for that entity, and disambiguation relies on the remaining signals $S_{sem}$ and $S_{sal}$.
%%%%%%%%%%%%%%%%%
The final set $\mathcal{K}$ is a domain-specific, user- and category-independent set of concepts. As stated earlier, although this set can be revised over time, we decided to keep it static for the work in this paper. In order to capture user-specific habits, the compositional hierarchy is extended to $\mathcal{L}_2$ by identifying workflow dependencies between these concepts.

%%%%%%%%%%%%%

\noindent
\paragraph{$\mathbf{\mathcal{L}_2}$\textbf{: User-specific attributes.}}
%The next level of the concept hierarchy seeks to identify user-specific \textit{habits} by applying statistics to mine the individual user's history of interacting with the agent. Specifically, for any given user, the agent filters $\mathcal{H}$ to retain only records of action execution (i.e., $hpd$ literals) that correspond to the user under consideration and match the current command under consideration (e.g., \emph{pickup}, \emph{switch\_on}). 
%%new added
The next level of the concept hierarchy seeks to identify user-specific \textit{habits} by applying heuristics to analyze any given human's history of interacting with the agent. Specifically, for any given user, the agent accesses $\mathcal{H}_u$, the history maintained separately for that user. It then extracts those records (i.e., $hpd$ literals) corresponding to an action that is the same as the current command. For instance, if the current command is \emph{switch\_on}, records of \emph{switch\_on} are extracted from the history of interactions with this user, while records of other commands are discarded.
%%%%%
This process ends up identifying a contextually matched subset of the personalized history. Each record in this subset is then examined to obtain an \emph{interaction snapshot} in the form of two heuristically modeled sets of related concepts of relevant objects.
\begin{itemize}

    % \item \emph{Trigger Concepts}: this set of concepts is defined as the union $\bigcup_{o \in O^{\Delta}_{t-1}} \Gamma(o)$. Recall that $O^{\Delta}_{t-1}$ is the set of entities that appear as an argument of a fluent literal whose value changed by the action executed by the agent at time $(t-1)$. Also, $\Gamma(o)$ is...;  and

    % \item \emph{Target Concepts}: this is the set of concepts $\Gamma(g_t)$ related to the entity selected at step $t$ in response to the current command. 
    \item \emph{Trigger concepts}: this is the set of concepts (i.e., combinations of attributes) of objects that were impacted by the commands similar to current command. Specifically, for each entity $o \in \mathcal{O}$, a set of objects of interest, let $\Gamma(o) \subseteq \mathcal{K}$ denote the set of $\mathcal{L}_1$ concepts describing $o$, i.e., those concepts whose constituent attributes are all contained in $o$'s attribute set $A(o)$. We can then describe the trigger concept set as: 
    \begin{align}
        \label{eqn:trigger}
        \bigcup_{o \in O^{\Delta}_{t-1}} \Gamma(o), \quad \mathrm{with} \quad \Gamma(o) = \{c \in \mathcal{K} \mid c \subseteq A(o)\}
    \end{align}
    where $O^{\Delta}_{t-1}$ is the set of entities that appear as an argument of a fluent literal whose value changed by the action executed by the agent at time $(t-1)$, and this change in value may have \textit{triggered} the execution of the action at time $t$. Note that this  trigger set is a simple union of the concepts of relevant objects.

    \smallskip
    \item \emph{Target concepts}: this is the set of $\mathcal{L}_1$ concepts $\Gamma(g_t)$ describing the entity $g_t$ (ground target object) that the agent acted on at step $t$ in the historical record under consideration. 
    
\end{itemize}
These snapshots capture the dependency between immediate world-state transitions and the subsequent user selections, particularly when the user engaged in similar interactions in the recent past. Across all snapshots, the agent enumerates trigger-target concept pairs and retains only those whose co-occurrence is both frequent and statistically significant, applying the same 'lift' criterion with a relaxed threshold $\lambda_2$ to accommodate the smaller sample size (i.e., limited number of examples) for the current user. The pairs of trigger concepts and target concepts that are retained form the user-specific workflow pattern set $P_u$. 

% For example, in the context of Figure~\ref{fig:compositional_hierarchy}, suppose a user's interaction history $\mathcal{H}$ contains multiple records of the execution of the action \emph{switch\_on}, and suppose that multiple entities in the domain can be switched on, e.g., table lamp, microwave, and ceiling fan. In one record, the preceding instruction ``Put a classic novel on the coffee table'' changed the state of the novel, coffee table, and bookshelf, and the union of the corresponding concepts (e.g., $c_{93}\{\textit{cuboid, paper, layered}\}, \ldots$) forms the trigger set. The user then interacted with the table lamp, whose concepts (e.g., $c_{21}\{\textit{hollow, cylinder, metallic}\}, \ldots$) form the target set. In another record, the trigger set derived from ``Bring me the magazine from desk'' again includes $c_{93}$, and the selected floor lamp again includes $c_{21}$. Across snapshots, the pair of concepts $(c_{93}, c_{21})$ co-occurs significantly more than chance ($\Pr(c_{93}, c_{21}) \gg \Pr(c_{93}) \cdot \Pr(c_{21})$) and is retained in $P_u$; other pairs such as $(c_{72}, c_{21})$ are not found to be very significant and are thus discarded.

For example, in the context of Figure~\ref{fig:compositional_hierarchy}, suppose a user's interaction history $\mathcal{H}$ contains multiple records of the execution of the action \emph{switch\_on}, and suppose that multiple entities in the domain can be switched on, e.g., the lamp, the chandelier, and the television. In one record, the preceding instruction ``Put the book on the table'' changed the state of the book, the table, and the bookshelf, and the union of the corresponding concepts (e.g., $c_{93}\{\textit{has\_printing\_on\_it, has\_printed\_text\_on\_the\_pages, has\_a\_cover}\}, \ldots$) forms the trigger set. The user then interacted with the lamp, whose concepts (e.g., $c_{21}\{\textit{emits\_light, contains\_lightbulbs, can\_be\_lit\_for\_light}\}, \ldots$) form the target set. In another record, the trigger set derived from ``Bring me the newspaper'' again includes $c_{93}$, and the selected chandelier again includes $c_{21}$. Across snapshots, the pair of concepts $(c_{93}, c_{21})$ co-occurs significantly more than chance, i.e., $\Pr(c_{93}, c_{21}) \gg \Pr(c_{93}) \cdot \Pr(c_{21})$, and is retained in $P_u$; other pairs such as $(c_{72}, c_{21})$ are not found to be very significant and are not included in $\mathcal{L}_2$ for this user.

%%%%%%%%%%%%%%%%%%%%%%%%%%%%%%%%%%%%%%%%%%%%%%%%%%%%%%%%%%%%%%%%%%%%%%%%%%%%%%%%
%%%%%%%%%%%%%%%%%%%%%%%%%%%%%%%%%%%%%%%%%%%%%%%%%%%%%%%%%%%%%%%%%%%%%%%%%%%%%%%%
\subsection{Ranking Ambiguous Candidates and Disambiguation}
\label{sec:problem-disambig}
Given the prior domain knowledge and the acquired compositional hierarchy, we return to the task of addressing ambiguity in the input command. The key part of Stage-2 of the pipeline of our architecture (in Figure~\ref{fig:framework}) is the filtering and ranking of candidate interpretations $\mathcal{C}_t$ of the input user command. Recall that multiple interpretations are possible when more than one domain object matches the under-specified reference in the input user command. Our architecture uses an approach comprising three steps for this filtering and ranking; the first step filters infeasible candidates based on prior domain knowledge, while the next two steps rank the remaining candidates. The ranking uses a computational model of contextual information that considers three complementary factors that are modeled and combined heuristically: semantic compatibility, session salience, and user-specific thematic preference, with the last factor leveraging the compositional hierarchy of concepts ($\mathcal{L}_1$ and $\mathcal{L}_2$ in Section~\ref{sec:problem-compositional}). 

\paragraph{\textbf{Step 1: Filtering.}}
As outlined in Section~\ref{sec:problem-kr}, the first step identifies candidates violating sort constraints in $\Sigma$ or executability conditions in $\Pi(\mathcal{D},\mathcal{H})$. These are filtered, i.e., removed from further consideration, because they are not valid choices given in the given domain. After this filtering, if $|\mathcal{C}_t|=0$, the agent requests clarification from the human; if $|\mathcal{C}_t|=1$, this single candidate is selected immediately as the unambiguous interpretation and used for subsequent steps (e.g., planning and plan execution).

\paragraph{\textbf{Step 2: Semantic similarity and salience.}}
When $|\mathcal{C}_t|>1$ after Step 1 above, each candidate $x \in \mathcal{C}_t$ is first scored by heuristic measures of semantic similarity and salience that consider recent contextual information but not the learned compositional hierarchy of attributes, i.e., information in $\mathcal{L}_1$ and $\mathcal{L}_2$ are not used. Specifically:
\begin{itemize}

    \item \textit{Semantic similarity} ($S_{sem}$): measures how well each candidate fits the context established by the preceding action, using a WordNet-based Wu--Palmer similarity score in $(0,1]$. To model this contextual information, the agent uses predefined associations between entities associated with the preceding action and the abstract context categories (e.g., \emph{dining}, \emph{reading}, \emph{relaxing}), weighted by association strength. Summing these weights across the entities in $O^{\Delta}_{t-1}$ and taking the highest-scoring category yields a single context keyword $w_{ctx}$ based on the preceding action. The Wu--Palmer similarity between each candidate $x \in \mathcal{C}_t$ and $w_{ctx}$ in WordNet then gives a score per candidate, which is normalized across $\mathcal{C}_t$ to sum to one.
    
    \smallskip
    \item \textit{Salience} ($S_{sal}$): measures how frequently each candidate has been the focus of attention in the recent past, within the current session. Specifically, each candidate $x \in \mathcal{C}_t$ accumulates geometrically decayed weights over the recent action history, with more recent occurrences contributing more, and the resulting scores are normalized across $\mathcal{C}_t$ to sum to one. 
\end{itemize}
 %consider 
% Let us consider, again, the example provided after the description of level $\mathcal{L}_2$ in Section~\ref{sec:problem-compositional}, in which the instruction ``Put a classic novel on the coffee table'' has just been executed. Now suppose the user issues an under-specified command ``switch it on,'' and Step~1 (above) leaves three candidates: \emph{table\_lamp}, \emph{microwave}, and \emph{ceiling\_fan}. Since the preceding action involves \textit{book-like} entities predominantly associated with the \emph{reading} context, $w_{ctx} = \emph{reading}$ is identified as the single context keyword. Then, since \emph{table\_lamp} is found to be closest to this context (\emph{reading}), it has the higher $S_{sem}$ score among the three candidates. Independently, the $S_{sal}$ measure favors entities most recently referenced in the current session; assuming that the lamp on the table was referenced in the recent past, we once again have \emph{table\_lamp} receive the highest $S_{sal}$ value among the three candidates. Based on both measures, \emph{table\_lamp} is ranked first among the three candidates based on the domain-specific measures.
Let us consider, again, the example provided after the description of level $\mathcal{L}_2$ in Section~\ref{sec:problem-compositional}, in which the instruction ``Put the book on the table'' has just been executed. Now suppose the user issues an under-specified command ``switch it on,'' and Step~1 (above) leaves three candidates: \emph{lamp}, \emph{microwave}, and \emph{television}. Since the preceding action involves \textit{book-like} entities predominantly associated with the \emph{reading} context, $w_{ctx} = \emph{reading}$ is identified as the single context keyword. Then, since \emph{lamp} is found to be closest to this context (\emph{reading}), it has the higher $S_{sem}$ score among the three candidates. Independently, the $S_{sal}$ measure favors entities most recently referenced in the current session; assuming that the lamp was referenced in the recent past, we once again have \emph{lamp} receive the highest $S_{sal}$ value among the three candidates. Based on both measures, \emph{lamp} is ranked first among the three candidates based on the domain-specific measures.

\paragraph{\textbf{Step 3: User-specific score.}}
%Human users interacting with the AI agent often exhibit distinct behavioral patterns. The thematic measure $S_{them}$ estimates personalized preferences using the matched records from the history of interaction of this user with the AI agent. These records are identified using the process described in Section~\ref{sec:problem-compositional} in the context of describing $\mathcal{L}_2$. It is a combination of two measures: $S_{action}(.)$ and $S_{pattern}(.)$. The first measure $S_{action}(x)$ scores each candidate by the frequency of its concepts $\Gamma(x)$ among entities in the matched subset. $S_{pattern}(x)$ scores each candidate by how well $\Gamma(x)$ completes the target side of patterns in $P_u$ activated by the current Trigger concepts. 
Human users interacting with the AI agent often exhibit distinct behavioral patterns. The thematic measure $S_{them}$ estimates personalized preferences using the compositional concept hierarchy representing each object and relevant records from the history of this user's interaction with the AI agent. These records are identified using the filtering process described in Section~\ref{sec:problem-compositional}. It combines, i.e., is the sum of, two heuristic measures that use the hierarchy of concepts:

\begin{itemize}
    \item \textit{Concept evidence} ($S_{concept}$): this gathers evidence of each candidate being the desired object, based on each $\mathcal{L}_1$ concept of the candidate and the user's recent interaction with objects with similar concepts using the action $a$ under consideration. Specifically, for a candidate $x$, the agent starts with a score of 0 and identifies each of its concepts $c \in \Gamma(x)$. It also considers all past records of action $a$ in the user's history ($\mathcal{H}_u$), and computes the proportion of these records that involve the concept $c$. This proportion is added to the score of candidate $x$, and the process is repeated for all entries in $\Gamma(x)$. 
    
    For example, if the user has habitually \emph{eaten} apples, apple-related concepts, e.g., \{\textit{sweet, crisp, can be red colored, ...}\}, will dominate the past \emph{eat} actions, so a candidate sharing more of them receives a higher $S_{concept}$ in the context of the eat action. Since this signal can cease to be useful when the same action is applied equally to different types of entities (e.g., \emph{switch\_on} for lamp, microwave, television), we also consider the measure below.

    \smallskip
    \item \textit{Pattern evidence} ($S_{pattern}$): this gathers evidence in support of each candidate $x$ based on set of $\mathcal{L}_2$ concepts for this user (i.e., $P_u$)---see Section~\ref{sec:problem-compositional}. Specifically, for each candidate $x$, the agent starts with a score of 0 and accumulates evidence based on each pattern $pt \in P_u$ whose target concept $c_{target}$ belongs to $\Gamma(x)$ and whose trigger concept $c_{trigger}$ is part of the previous command. This evidence is the product of two factors: \emph{follow rate} and \emph{lift}, where $\mathrm{lift}(c_{trigger}, c_{target})$ is computed as in Equation~\ref{eqn:lift}, with probabilities $\Pr(c_{trigger}, c_{target})$, $\Pr(c_{trigger})$ and $\Pr(c_{target})$  estimated from this user's history $\mathcal{H}_u$.
    %This evidence is the product of two factors: \emph{follow rate} and \emph{lift}. 
    The follow rate considers past records (for this user) in  which the trigger concept appeared at step $t-1$, and computes the fraction of these records in which the target concept appeared at step $t$. 
    
    For instance, if a book-like concept appeared at 20 times (in this user's history) and was followed by a lamp-like concept 12 times, the follow rate is $12/20=0.6$. Follow rate, by itself, can overemphasize the evidence from common targets, but lift (computed as above) can substantially emphasize small-sample coincidences. So the net evidence accumulated by $S_{pattern}$ is the product of follow rate and \emph{lift}. Continuing with our example, when book-like and table-like concepts both appear in $O^{\Delta}_{t-1}$, the agent looks up $P_u$ and finds two matched patterns, both with a lamp-like target concept and evidence (computed as described above) of $0.55$ and $0.30$ respectively. $\Gamma(x)$ for the lamp candidate contains the lamp-like concept, the target of both patterns, so the candidate accumulates a value of $0.55 + 0.30 = 0.85$. $\Gamma(x)$ for the other two candidates, microwave and television, does not contain this concept; they accumulate a value of $0$. 
    
    %which the trigger concept appeared at step $t-1$, the fraction in which the target concept appeared at step $t$. For instance, if a book-like concept appeared at step $t-1$ 20 times in this user's history and was followed by a lamp-like concept at step $t$ in 12 of them, the follow rate is $12/20=0.6$. The second factor is $\log(\text{lift})$, which captures how strongly the trigger and target are associated beyond chance. Follow rate alone over-rewards common targets, and $\log(\text{lift})$ alone over-rewards small-sample coincidences; the product is trying to avoids both. Continuing the example, when book-like and coffee\_table-like concepts both appear in $O^{\Delta}_{t-1}$, the agent looks up $P_u$ and finds two matched patterns, both with a lamp-like target concept and strengths $0.55$ and $0.30$ respectively. $\Gamma(x)$ for the lamp candidate contains the lamp-like concept, the target of both patterns, so the candidate accumulates $0.55 + 0.30 = 0.85$. $\Gamma(x)$ for the microwave and ceiling fan candidates does not contain this concept, so they accumulate $0$. Together with $S_{concept}$, this provides complementary evidence that the agent combines into $S_{them}$.
\end{itemize}
%%%%
The thematic score then computed as the sum of the individual scores:
\begin{equation}
S_{them}(x) = S_{concept}(x) + S_{pattern}(x).
\end{equation}
Note that this combined score models the preference of the specific user based on their personalized history.

\paragraph{\textbf{Combined Score and Selection.}}
$S_{sem}$ is a similarity measure, whereas $S_{sal}$ and $S_{them}$ are essentially counts (sum or fractions). In order to combine these measure, we first make them comparable by normalizing each measure to obtain a value $\in [0, 1]$ over the set of candidates ($\mathcal{C}_t$). The individual measures are then fused by computing their weighted sum:
\begin{equation}
\label{eqn:score_fusion}
% Score(x) \;=\; S_{sem}(x) \;+\; S_{sal}(x) \;+\; S_{them}(x).
Score(x) \;=\; w_{sem}\,S_{sem}(x) \;+\; w_{sal}\,S_{sal}(x) \;+\; w_{them}\,S_{them}(x),
\end{equation}
where the weights are experimentally determined from training data for each ambiguity level and shared by all methods that compute a match score. These weights can be viewed as the relative \textit{validity} of each of the scores being combined. The normalized signals still differ in their values and purpose, e.g., salience concentrates on the recently mentioned candidate(s) whereas the thematic signal assigns similar weights to candidates that any particular user has interacted with. As discussed in Section~\ref{sec:exp-setup}, the observed experimental results are not sensitive to the choice of weights over a range of values.

After all factors of contextual information are considered in ranking the available candidates, the agent selects the top-ranked candidate only when its margin over the candidate ranked second is significant. If this is not the case, the agent requests clarification from the human. Once the disambiguation is eventually achieved, and the agent ends up executing actions to complete the tasks, history $\mathcal{H}$ is updated with the new record and used for disambiguation in the subsequent interactions.

%%%%%%%%%%%%%%%%%%%%%%%%%%%%%%%%%%%%%%%%%%%%%%%%%%%%%%%%%%%%%%%%%%%%%%%%%%%%%%%%

\section{Experimental Setup and Results}
\label{sec:exp-setup-results}
We experimentally evaluated five hypotheses regarding our architecture's capabilities:
\begin{itemize}
\item[\textbf{H1:}] Integrated reasoning and learning with the compositional hierarchy enables transfer of knowledge of shared concepts from previously-observed entities to entities with limited (or no) interaction history.

\item[\textbf{H2:}] Our architecture captures personalized preferences more effectively than LLM baselines, both with and without human-in-the-loop clarification, with the benefits increasing in more personalized scenarios.

\item[\textbf{H3:}] Aggregating statistics at the concept level ($\mathcal{L}_1$) improves disambiguation compared with statistics at the level of individual attributes ($\mathcal{L}_0$).
%, and both outperform entity-name-level modeling

\item[\textbf{H4:}] Semantic compatibility, session salience, and user-specific thematic preference provide complementary evidence; their combination outperforms any factor considered individually.

\item[\textbf{H5:}] Concept-level statistics ($\mathcal{L}_1$) suffice under low levels of ambiguity, but user-specific patterns ($\mathcal{L}_2$) become more useful as ambiguity increases.

% \item[\textbf{H5:}] The proposed architecture captures individual user habits more effectively than LLM baselines, with the advantage increasing in highly personalized scenarios.

\end{itemize}
%%%%%%%%%%%%%%%%%%%%%%%%%%%%%%%%%%%%%%%%%%%%%%%%%%%%%%%%%%%%%%%%%%%%%%%%%%%%%%%%
\subsection{Experimental Setup}
\label{sec:exp-setup}

% We created a household environment with $42$ entities, four rooms, and $11$ actions. Each entity was decomposed into atomic attributes across three dimensions (structure, shape, physical properties) using GPT-5.1~\cite{singh2025openai} as a structured annotation tool, yielding an attribute vocabulary of $|\mathcal{A}| = 106$. We mined $\mathcal{L}_1$ concepts by retaining attribute combinations of size $2$--$5$ that co-occur across at least $10\%$ of entities and satisfy $\lambda_1 = 1.3$, producing $204$ concepts effectively mapped to entities. For $\mathcal{L}_2$ workflow pattern mining, we set $\lambda_2 = 1.2$.

For experimental evaluation, we created a household environment with $66$ entities, four rooms, and $12$ actions. Each entity is represented by its atomic attributes from the NOVA semantic feature norms dataset~\cite{suresh2025nova} (Section~\ref{sec:problem-compositional}), yielding an attribute vocabulary of $|\mathcal{A}| = 3{,}983$ across the $66$ entities. We mined $\mathcal{L}_1$ concepts by retaining attribute combinations of size $2$--$5$ that co-occur in at least $10\%$ of the entities and satisfy the lift threshold $\lambda_1 = 2.0$, producing $2{,}508$ concepts mapped to the $66$ entities. For $\mathcal{L}_2$ workflow pattern mining, we set $\lambda_2 = 1.2$ with a minimum support of two snapshots; each user's history yields $138$--$150$ interaction snapshots from which these patterns are mined.
%new Added 11/6
% Since triggers and targets are concept sets, these pairs are enumerated directly from the snapshots rather than stored as explicit rules, and only the pairs linking the previous step's trigger concepts to the current candidates' concepts are consulted.

The command set contains $200$ high-level commands (e.g., rearranging objects in specific configurations, fetching specific objects), each annotated with a textual description and an ASP-based goal, e.g., ``has(user, book)'' or ``on(cup, table)''. As with other such simulated domains, the agent has knowledge of the domain's state (i.e., full observability). To capture realistic variation in user habits, five human participants not involved in our architecture's design were recruited to provide descriptions of their household routines and preferences. Based on their input, we constructed and verified $10$ interaction sessions per user, each containing $16$ sequential commands under different initial conditions. For any given user, the $10$ sessions were split so that one session was held out for testing and the remaining nine were used as historical information, with the held-out session excluded when computing $S_{them}$. We ran ten such trials per user, each holding out a different session, across all five users. Any given session comprises a coherent multi-step routine (e.g., coffee preparation, relaxing in the living room, studying, cooking). Sessions involving the same user share consistent preferences, while sessions across users exhibit systematically different preferences to ensure that user-specific history is useful for accurate disambiguation. In three sessions per user, the target objects are novel, appearing in no other session and coming from categories adjacent to familiar ones, e.g., a mug where the user's routines feature cups, or an apricot where they feature peaches. When such a session is held out for testing, these objects have no direct history of their own, so resolving them depends on transferring any given user's preferences related to the concepts that the novel object shared with the objects familiar to the AI agent.

%, yielding $3{,}000$ ambiguous commands evaluated under both \emph{noask} and \emph{ask} modes.

Recall that this paper focuses on exploring the use of compositionality and heuristics in the design of architectures for assistive AI agents. The \textbf{performance task} of an assistive AI agent addressing ambiguity in the object(s) being referred to in the input commands (from a human) is just the representative use case considered in this paper. Ambiguity was generated by systematically replacing explicit object mentions with increasingly ambiguous references at four levels: 
\begin{itemize}
    \item[\textbf{A1}] Near hypernym (e.g., ``apple''$\rightarrow$``fruit'')
    \item[\textbf{A2}] Higher hypernym (e.g., ``apple''$\rightarrow$``food'')
    \item[\textbf{A3}] Broad hypernym (e.g., ``apple''$\rightarrow$``object'')
    \item[\textbf{A4}] Pronoun (e.g., ``apple''$\rightarrow$``it/that'')
\end{itemize}
%%%%%new add for the number of candidate
% The average number of candidates per command is $3.5$, $14.6$, $29.0$, and $42.0$ (out of $42$ entities) for A1--A4 respectively.
The average number of candidates per command is $6.0$, $16.6$, $52.2$, and $66.0$ (out of $66$ entities) for A1--A4 respectively. This is because the four levels weaken the evidence available for disambiguation to different extents: the near hypernym (A1) still constrains the entity type, whereas the pronoun (A4) carries no type information and resolution must rely almost entirely on the session context and the user's habits.

%%%%%
For each session, we keep the first command unambiguous and apply ambiguities to the remaining commands, yielding $3{,}000$ ambiguous commands in total ($5$ users $\times 10$ sessions $\times 15$ commands $\times 4$ levels). Commands in each session are processed sequentially. After each command, the symbolic world state is updated with the ground-truth referent to isolate errors in subsequent steps. For evaluation, each test session uses the remaining $9$ sessions of the same user as history. $\mathcal{L}_2$ workflow patterns are mined from records in these sessions whose action predicate matches the current instruction, as described in Section~\ref{sec:problem-compositional}.

We report results under two modes: \emph{noask}, where a method must always commit to a selection (the one with top score), and \emph{ask}, where it may request clarification when evidence is insufficient. For LLM baselines under \emph{ask} mode, the model is allowed to output a clarification request when uncertain; under \emph{noask} mode, clarification is not allowed and the model is forced to commit to a candidate. We used GPT-5.1 as the LLM model in all the baselines that include an LLM. For the proposed method, clarification is triggered when the top candidate's lead over the runner-up falls below a threshold (see Section~\ref{sec:problem-disambig}).
% we use fixed thresholds per ambiguity level, set empirically: $0.18$ for A1, $0.16$ for A2, and $0.15$ for both A3 and A4. 
This threshold is set to be $0.25$ at all ambiguity levels based on the observation that once all the scores have been computed and merged, there are $\le 3-4$ good candidates with a non-trivial score. 

All methods that compute a match score (Equation~\ref{eqn:score_fusion} in Section~\ref{sec:problem-disambig}), including the ablations, share the same fusion weights.
% Specifically, we experimentally set $(w_{sem}, w_{them}, w_{sal})$ to $(0.40, 0.45, 0.15)$ for A1, $(0.25, 0.62, 0.13)$ for A2 and A3, and $(0.35, 0.52, 0.13)$ for A4, giving the semantic signal more weight at A1, where the hypernym still constrains the candidate type. 
As stated earlier, these weights associated with the individual heuristically-modeled measures (factors) were computed experimentally. Specifically, we used the relative contribution of the individual methods to the known (correct) outcome in the training set to estimate these weights. Since different sessions are considered for the training set and  the held-out testing set in each trial, the estimated values of the weights could differ. However, we observed some common trends. For example, the thematic measure always received the largest weight, and the weights were always within a narrow band: $w_{them} \in [0.45, 0.62]$ and $w_{sal} \in [0.13, 0.15]$, with $w_{sem}$ taking the remainder. In addition, we observed that the performance was not very sensitive to the value of the weights as along as relative values of the weights were similar; the average accuracy (in the disambiguation task) varied by $\le 1\%$ across a range of these values such as $w_{them} \in [0.55, 0.80]$, $w_{sal} \in [0.10, 0.15]$.

%%%%%%%%%%%%%%%%%%%%%%%%%%%%%%%%%%%%%%%%%%%%%%%%%%%%%%%%%%%%%%%%%%%%%%%%%%%%%%%%
\subsection{Baselines and Performance Measures} 
\label{subsec:baselines}
We experimentally evaluated and compared the performance of our architecture with eight baselines:
\begin{enumerate}
    \item[\textbf{(B0)}] \textbf{Random.} This strategy selects one of the candidates randomly from the set obtained after the ASP-based filtering (in Stage 1, as described in Section~\ref{sec:problem-kr}).
    
    \item[\textbf{(B1)}] \textbf{LLM full-information + CoT.}  In this strategy, GPT-5.1 seeks to resolve the ambiguity by selecting one candidate from the full set of potential candidates using chain-of-thought prompting~\cite{wei2022chain}, based on the current state of the domain, the current session, and the user's history of interactions that is also available to our architecture.
    % GPT-5.1 receives the ambiguous command, the full (unfiltered) candidate list, the complete symbolic world state (the current location and state of every entity in the domain), the preceding commands of the current session, and the user's cross-session interaction history (the same history available to our method); it selects one candidate using chain-of-thought prompting.
    
    \item[\textbf{(B2)}] \textbf{LLM full-information + CoT, ASP-based filtering.} This strategy builds on \textbf{B1}, with ASP-based feasibility filtering applied first before the LLM selects among the remaining feasible candidates. The LLM thus receives the inputs available to our architecture after the ASP-based filtering.
    
    \item[\textbf{(B3)}] \textbf{Semantic and salience scores, ASP-based filtering.} In this strategy, ASP-based feasibility filtering first eliminates some candidates. Then the $S_{sem}$ and $S_{sal}$ scores are combined using the same method (Equation~\ref{eqn:score_fusion}) and thresholds used in our architecture, but without considering the contribution of the user-specific thematic preference score ($S_{them}$).
    
    \item[\textbf{(B4)}] \textbf{Object-level history, contextual scores, ASP-based filtering, no compositional concepts.} This strategy uses the ASP-based filtering and the combination of the three scores included in our architecture, but thematic preference is computed over object categories instead of the hierarchical concepts. As a result, this strategy explores the contribution of the compositional concepts.
    
    \item[\textbf{(B5)}] \textbf{$\mathcal{L}_0$ attributes, ASP-based filtering.} This strategy builds on the strategy for \textbf{B4}, with the difference being that the thematic preference score is computed over the raw $\mathcal{L}_0$ atomic attributes of each object/entity rather than over object categories. As a result, this strategy further explores the contribution of concepts.
    
    \item[\textbf{(B6)}] \textbf{$\mathcal{L}_1$ concepts, ASP-based filtering.} This strategy builds on the strategy for \textbf{B4}, i.e., candidates that remain after ASP-based filtering are ranked based on the three factors. The difference is that the thematic score $S_{them}$ is computed using concept-level evidence ($S_{concept}$ based on $\mathcal{L}_1$) but without the user-specific patterns ($S_{pattern}$ based on $\mathcal{L}_2$), examining the contribution of user-specific workflow patterns.
    
    \item[\textbf{(B7)}] \textbf{Full system, no ASP-based filtering.} This strategy is very similar to our architecture in the use of the three factors and the compositional hierarchy for scoring the potential candidates for disambiguation, but without the ASP-based initial filtering of the candidates. As a result, this examines the contribution of logical reasoning with prior knowledge.
\end{enumerate}

\noindent
Compared with these baselines, \textbf{our architecture} ("proposed") uses ASP-based filtering to remove physically infeasible candidates, and scores the remaining candidates by fusing three complementary signals: semantic similarity ($S_{sem}$), session salience ($S_{sal}$), and user-specific thematic preference ($S_{them}$) computed through compositional hierarchy of concepts (in $\mathcal{L}_0$, $\mathcal{L}_1$, $\mathcal{L}_2$). It also requests clarification from a human only when it is not able to identify a clear candidate that addresses the ambiguity.

For paired comparison with baselines, each specific experimental run with our architecture was repeated for each baseline with the same users, sessions, ambiguous command instances, candidate sets, and history splits.
%%%%%%%%%%%%%%%%%%%%%%%%%%%%%%%%%%%%%%%%%%%%%%%%%%%%%%%%%%%%%%%%%%%%%%%%%%%%%%%%
% \subsection{Evaluation Measures}
% \label{sec:measures}
We used three \textbf{performance measures} in our experiments:
\begin{enumerate}
    \item \textbf{Overall accuracy.} This is computed as the proportion of commands for which the correct object was selected to address the ambiguity; in the \emph{ask} mode, soliciting clarification from a human is considered as an incorrect outcome.
    
    \item \textbf{Answer accuracy.} This is similar to the previous measure but computed only over commands where the method made a definite selection, i.e., trials with clarifications are excluded.
    
    \item \textbf{Clarification rate.} This is computed as the proportion of commands for which the approach being evaluated required clarification from a human to address the ambiguity.
\end{enumerate}

%%%%%%%%%%%%%%%%%%%%%%%%%%%%%%%%%%%%%%%%%%%%%%%%%%%%%%%%%%%%%%%%%%%%%%%%%%%%%%%%
%%%%%%%%%%%%%%%%%%noask%%%%%%%%%%%%%%%%%%%%%%%%%%%
%%%% Table 1
\begin{table}[tb]
\caption{\textnormal{Comparison of our architecture with \textbf{B0--B7} under \emph{noask} mode; ($\uparrow$) indicates larger values are better, and numbers in boldface indicate that the corresponding method is significantly better ($99\%$ level of significance) than the others. Our architecture outperforms all baselines at every ambiguity level. The gap between \textbf{B6} and our architecture is largest at A3--A4, indicating that workflow patterns become more useful as ambiguity increases; the LLM-only baselines (\textbf{B1--B2}) achieve substantially lower accuracy despite receiving the same interaction history.} }
\label{tab:bylevel-noask}
\vspace{-1em}
\centering
\scriptsize
\renewcommand{\arraystretch}{1.4}
\resizebox{\textwidth}{!}{%
\begin{tabular}{|l|c|c|c|c|c|c|c|c|c|}
\hline
\textbf{Ambiguity Level} $\rightarrow$
& \textbf{\begin{tabular}[c]{@{}c@{}}B0:\\ Random\end{tabular}}
& \textbf{\begin{tabular}[c]{@{}c@{}}B1:\\ LLM+CoT\end{tabular}}
& \textbf{\begin{tabular}[c]{@{}c@{}}B2:\\ +CoT+ASP\end{tabular}}
& \textbf{\begin{tabular}[c]{@{}c@{}}B3:\\ $S_{sem}$+$S_{sal}$\end{tabular}}
& \textbf{\begin{tabular}[c]{@{}c@{}}B4:\\ Object-level\end{tabular}}
& \textbf{\begin{tabular}[c]{@{}c@{}}B5:\\ $\mathcal{L}_0$ attr.\end{tabular}}
& \textbf{\begin{tabular}[c]{@{}c@{}}B6:\\ Concept-level\end{tabular}}
& \textbf{\begin{tabular}[c]{@{}c@{}}B7:\\ No ASP\end{tabular}}
& \textbf{Proposed} \\
\hline
\textbf{Accuracy A1 (\%) $\uparrow$} & 27.6 & 46.5 & 54.0 & 39.1 & 38.7 & 46.2 & 63.1 & 62.2 & \textbf{74.2} \\
\hline
\textbf{Accuracy A2 (\%) $\uparrow$} & 7.1 & 33.2 & 40.4 & 26.2 & 25.8 & 28.4 & 52.0 & 52.9 & \textbf{64.4} \\
\hline
\textbf{Accuracy A3 (\%) $\uparrow$} & 2.2 & 23.1 & 26.3 & 20.9 & 19.6 & 21.8 & 36.4 & 45.8 & \textbf{57.3} \\
\hline
\textbf{Accuracy A4 (\%) $\uparrow$} & 1.3 & 20.9 & 22.9 & 19.6 & 18.7 & 20.9 & 35.6 & 35.1 & \textbf{54.7} \\
\hline
\textbf{Average (\%) $\uparrow$} & 9.6 & 30.9 & 35.9 & 26.4 & 25.7 & 29.3 & 46.8 & 49.0 & \textbf{62.7} \\
\hline
\end{tabular}%
}
\end{table}

%%%%%%%%%%%%%%%%%%ask%%%%%%%%%%%%%%%%%%%%%%%%%%%
%%%% Table 2
\begin{table}[t]
\caption{\textnormal{Comparison of our architecture with \textbf{B0--B7} under \emph{ask} mode; ($\uparrow$) implies larger values are better, ($\downarrow$) implies that smaller values are better, and numbers in boldface indicate that the corresponding method is significantly better ($99\%$ level of significance) than the others. The LLM baselines without the compositional hierarchy or heuristics (\textbf{B1--B2}) solicit clarifications from a human much more than our architecture but are still less accurate in addressing ambiguity. Our architecture solicits clarifications the least, is the most accurate when it commits, and attains the highest overall accuracy at every ambiguity level.} }
\label{tab:bylevel-ask}
\vspace{-1em}
\centering
\scriptsize
\renewcommand{\arraystretch}{1.4}
\resizebox{\textwidth}{!}{%
\begin{tabular}{|l|c|c|c|c|c|c|c|c|c|}
\hline
\textbf{Ambiguity Level} $\rightarrow$
& \textbf{\begin{tabular}[c]{@{}c@{}}B0:\\ Random\end{tabular}}
& \textbf{\begin{tabular}[c]{@{}c@{}}B1:\\ LLM+CoT\end{tabular}}
& \textbf{\begin{tabular}[c]{@{}c@{}}B2:\\ +CoT+ASP\end{tabular}}
& \textbf{\begin{tabular}[c]{@{}c@{}}B3:\\ $S_{sem}$+$S_{sal}$\end{tabular}}
& \textbf{\begin{tabular}[c]{@{}c@{}}B4:\\ Object-level\end{tabular}}
& \textbf{\begin{tabular}[c]{@{}c@{}}B5:\\ $\mathcal{L}_0$ attr.\end{tabular}}
& \textbf{\begin{tabular}[c]{@{}c@{}}B6:\\ Concept-level\end{tabular}}
& \textbf{\begin{tabular}[c]{@{}c@{}}B7:\\ No ASP\end{tabular}}
& \textbf{Proposed} \\
\hline
\multicolumn{10}{|l|}{\textbf{Overall accuracy (\%)} \textit{(a clarification counts as not completed)}} \\
\hline
\textbf{A1 $\uparrow$} & 27.6 & 25.9 & 29.7 & 32.0 & 27.1 & 28.9 & 57.3 & 54.2 & \textbf{66.7} \\
\hline
\textbf{A2 $\uparrow$} & 7.1 & 19.7 & 22.5 & 24.4 & 19.1 & 20.9 & 38.2 & 43.6 & \textbf{54.2} \\
\hline
\textbf{A3 $\uparrow$} & 2.2 & 18.1 & 19.7 & 13.3 & 11.6 & 12.9 & 27.6 & 36.0 & \textbf{47.6} \\
\hline
\textbf{A4 $\uparrow$} & 1.3 & 16.0 & 18.8 & 12.0 & 10.7 & 12.0 & 26.7 & 26.7 & \textbf{44.4} \\
\hline
\textbf{Average $\uparrow$} & 9.6 & 19.9 & 22.7 & 20.4 & 17.1 & 18.7 & 37.4 & 40.1 & \textbf{53.2} \\
\hline
\multicolumn{10}{|l|}{\textbf{Answered accuracy (\%)} \textit{(over the commands each method chose to answer)}} \\
\hline
\textbf{A1 $\uparrow$} & 27.6 & 64.5 & 70.1 & 49.8 & 49.4 & 50.3 & 80.7 & 75.4 & \textbf{82.6} \\
\hline
\textbf{A2 $\uparrow$} & 7.1 & 49.3 & 54.7 & 35.6 & 33.2 & 35.9 & 69.3 & 63.6 & \textbf{72.7} \\
\hline
\textbf{A3 $\uparrow$} & 2.2 & 27.2 & 31.2 & 29.7 & 27.8 & 29.6 & 58.6 & 55.6 & \textbf{69.0} \\
\hline
\textbf{A4 $\uparrow$} & 1.3 & 23.4 & 27.4 & 28.6 & 26.3 & 28.1 & 54.1 & 44.7 & \textbf{66.3} \\
\hline
\textbf{Average $\uparrow$} & 9.6 & 41.1 & 45.9 & 35.9 & 34.2 & 35.9 & 65.7 & 59.8 & \textbf{72.7} \\
\hline
\textbf{Clarification (\%) $\downarrow$}
& -- & 46.2 & 46.1 & 46.1 & 56.6 & 54.7 & 44.2 & 33.9 & \textbf{26.6} \\
\hline
\end{tabular}%
}
\end{table}

%%%%%%%%%%%%%%%%%%%%%%%%%%%%%%%%%%%%%%%%%%%%%%%%%%%%%%%%%%%%%%%%%%%
\subsection{Experimental Results}
\label{subsec:results}
% In the experiments, the $20$ sessions for any given user were evaluated by holding out one session for testing and using the remaining $19$ sessions as user-specific history. We repeated this for all $20$ sessions across all five users, yielding $2{,}800$ ambiguous commands evaluated under both \emph{noask} and \emph{ask} modes.

% For any given user, the $10$ sessions were split so that one session was held out for testing and the remaining nine were used as historical information, with the held-out session excluded when computing $S_{them}$. We ran ten such trials per user, each holding out a different session, across all five users, yielding $3{,}000$ ambiguous commands evaluated under both \emph{noask} and \emph{ask} modes.

We now summarize and discuss the results of experimental evaluation.

\paragraph{\textbf{Overall comparison.}}
Tables~\ref{tab:bylevel-noask} and~\ref{tab:bylevel-ask} summarize the results comparing our ("proposed") architecture with \textbf{B0--B7}. Our architecture led to significantly higher accuracy than all baselines at every ambiguity level (A1-A4) under both modes (\textit{noask}, \textit{ask}). We also observe the following:
\begin{itemize}
    \item Performance degrades the least with our architecture as ambiguity increases from A1 to A4. For example, the accuracy with our architecture drops from $74.2\%$ to $54.7\%$ in Table~\ref{tab:bylevel-noask} as ambiguity increases from A1 to A4, but is still much better than any of the baselines; similar performance is observed in Table~\ref{tab:bylevel-ask}.

    \item The LLM-only baselines that do not use the compositional hierarchy or the simple heuristic measures (\textbf{B1--B2}) achieve substantially lower accuracy than our architecture despite receiving the same interaction history. This result indicates the importance of the compositional hierarchy and the heuristics.

    \item The ASP-based filtering (i.e., reasoning logically with prior domain knowledge to eliminate incorrect candidates) plays an important role in our architecture's performance; removing this component (e.g., in \textbf{B7}) reduced the accuracy substantially in Tables~\ref{tab:bylevel-noask} and~\ref{tab:bylevel-ask}.

    \item Each layer of the compositional hierarchy contributes to our architecture's performance, e.g., accuracy drops substantially with increasing ambiguity without this hierarchy. Also, the performance gap between \textbf{B6} and our architecture is largest at A3--A4, i.e., workflow patterns are more useful as ambiguity increases.
\end{itemize}
These results jointly provide evidence in support of all hypotheses \textbf{H1-H5}. The results of experiments described below explore these hypotheses further.

%%%%%%%%%%%%%%%%%% freq_transfer %%%%%%%%%%%%%%%%%%
%%% Table 3
\begin{table}[t]
\caption{\textnormal{Accuracy (\%) under \emph{noask} mode, grouped by how often the command's target entity occurs in the user's history; ($\uparrow$) implies larger values are better, and numbers in boldface indicate that the corresponding methods is significantly better (99\% level of significance) than others. \textbf{B4} is effective only with frequent targets, while our architecture provides good accuracy even for rare targets. } }
\label{tab:freq-transfer}
\vspace{-1em}
\centering
\small
\setlength{\tabcolsep}{5pt}
\begin{tabular}{@{}lrrr@{}}
\toprule
\textbf{Method}
& \textbf{Rare ($\leq 5$) $\uparrow$}
& \textbf{Medium ($6$--$10$) $\uparrow$}
& \textbf{Frequent ($11$--$20$) $\uparrow$} \\
\midrule
B4: Object-level & 16.2 & 19.6 & 50.6 \\
B6: Concept-level (no pattern) & 24.2 & 41.8 & 83.2 \\
Proposed & \textbf{44.6} & \textbf{60.8} & \textbf{89.0} \\
\bottomrule
\end{tabular}
\end{table}

%%% Table 4
\begin{table}[t]
\caption{\textnormal{Accuracy of our architecture with different sources of history of users' interaction under \emph{noask} mode. As before, ($\uparrow$) indicates larger values are better, and numbers in boldface indicate that the corresponding method is significantly better than others (99\% level of significance). The different options only change the history used to compute $S_{them}$. In \emph{mismatched user} setting, each user is paired with a different user's history.} }
\label{tab:matched-mismatched}
\vspace{-1em}
\centering
\small
\setlength{\tabcolsep}{5pt}
\begin{tabular}{@{}lrrrrr@{}}
\toprule
\textbf{History Source}
& \textbf{A1 $\uparrow$}
& \textbf{A2 $\uparrow$}
& \textbf{A3 $\uparrow$}
& \textbf{A4 $\uparrow$}
& \textbf{Avg. $\uparrow$} \\
\midrule
Matched user & \textbf{74.2} & \textbf{64.4} & \textbf{57.3} & \textbf{54.7} & \textbf{62.7} \\
Mismatched user & 31.2 & 15.5 & 8.8 & 8.3 & 16.0 \\
No user history (B3) & 39.1 & 26.2 & 20.9 & 19.6 & 26.4 \\
\bottomrule
\end{tabular}
\end{table}

%%%%%%%%%%%%%%%% history-depth + tradeoff %%%%%%%%%%%%%%%%
\begin{figure}[t]
\centering
\begin{minipage}[t]{0.49\textwidth}
\centering
\textbf{(a)}\par\vspace{-5pt}
\begin{tikzpicture}[baseline=(current axis.north west)]
\begin{axis}[
    width=\linewidth, height=0.72\linewidth,
    xlabel={Number of historical sessions}, ylabel={Average accuracy (\%)},
    xmin=-0.3, xmax=9.3, ymin=10, ymax=70,
    xtick={0,2,4,6,8}, ytick={10,20,30,40,50,60,70},
    grid=major,
    label style={font=\scriptsize}, tick label style={font=\scriptsize},
    legend style={font=\tiny, draw=none, at={(0.5,-0.30)}, anchor=north, legend columns=2},
    legend cell align=left,
    /tikz/every even column/.append style={column sep=4pt},
]
\addplot[color=magenta, mark=+, mark size=2.2pt, line width=1pt]
    coordinates {(0,24.2)(1,40.2)(2,44.4)(3,44.4)(4,45.6)(5,46.5)(6,46.6)(7,46.1)(8,45.7)(9,46.1)};
\addlegendentry{B6 (concept, no pattern)}
\addplot[color=blue, mark=square, mark size=2pt, line width=1pt]
    coordinates {(0,22.6)(1,25.1)(2,28.8)(3,29.2)(4,29.8)(5,31.2)(6,31.7)(7,31.3)(8,31.5)(9,30.9)};
\addlegendentry{B1 (LLM)}
\addplot[color=orange!90!black, mark=triangle, mark size=2.2pt, line width=1pt]
    coordinates {(0,19.2)(1,22.2)(2,23.4)(3,24.7)(4,25.3)(5,25.3)(6,25.9)(7,26.3)(8,26.3)(9,26.1)};
\addlegendentry{B4 (object)}
\addplot[color=black, mark=triangle*, mark size=2.5pt, line width=1.5pt]
    coordinates {(0,24.2)(1,40.2)(2,45.6)(3,51.6)(4,55.1)(5,58.0)(6,59.6)(7,61.1)(8,62.8)(9,63.3)};
\addlegendentry{Proposed}
\end{axis}
\end{tikzpicture}
\end{minipage}
\hfill
\begin{minipage}[t]{0.49\textwidth}
\centering
\textbf{(b)}\par\vspace{-5pt}
\begin{tikzpicture}[baseline=(current axis.north west)]
\begin{axis}[
    width=\linewidth, height=0.72\linewidth,
    xlabel={Clarification Rate (\%)}, ylabel={Overall Accuracy (\%)},
    xmin=-5, xmax=65, ymin=0, ymax=65,
    grid=major,
    label style={font=\scriptsize}, tick label style={font=\scriptsize},
    legend style={font=\tiny, draw=none, at={(0.5,-0.30)}, anchor=north, legend columns=5},
    legend cell align=left,
    /tikz/every even column/.append style={column sep=4pt},
]
\addplot[only marks, mark=x, mark size=2.5pt, color=brown, line width=1pt] coordinates {(0.00,9.6)};
\addlegendentry{B0}
\addplot[only marks, mark=square, mark size=2pt, color=blue, line width=1pt] coordinates {(46.2,19.9)};
\addlegendentry{B1}
\addplot[only marks, mark=diamond, mark size=2.5pt, color=red, line width=1pt] coordinates {(46.1,22.7)};
\addlegendentry{B2}
\addplot[only marks, mark=o, mark size=2pt, color=green!60!black, line width=1pt] coordinates {(46.1,20.4)};
\addlegendentry{B3}
\addplot[only marks, mark=triangle*, mark size=3pt, color=black, line width=1pt] coordinates {(26.6,53.2)};
\addlegendentry{Proposed}
\addplot[only marks, mark=triangle, mark size=2.5pt, color=orange!90!black, line width=1pt] coordinates {(56.6,17.1)};
\addlegendentry{B4}
\addplot[only marks, mark=otimes, mark size=2.3pt, color=violet, line width=1pt] coordinates {(54.7,18.7)};
\addlegendentry{B5}
\addplot[only marks, mark=+, mark size=2.5pt, color=magenta, line width=1.5pt] coordinates {(44.2,37.4)};
\addlegendentry{B6}
\addplot[only marks, mark=star, mark size=2.5pt, color=cyan!70!black, line width=1pt] coordinates {(33.9,40.1)};
\addlegendentry{B7}
\end{axis}
\end{tikzpicture}
\end{minipage}
\vspace{-1em}
\caption{\textnormal{(a) Accuracy in \emph{noask} mode (averaged over the four ambiguity levels) as a function of the number of historical sessions used to compute the user's patterns; accuracy with our architecture is substantially higher compared with not using the user-specific patterns and/or the compositional hierarchy; (b) Tradeoff between clarification and accuracy in the \emph{ask} mode; the ideal point is the upper-left corner, indicating high accuracy while requiring minimal clarification. Our architecture gets us closest to this ideal point, much better than the LLM baselines (\textbf{B1--B2}) and any combination of not using the compositional hierarchy or the heuristic measures. }}
\label{fig:tradeoff-history}
\end{figure}
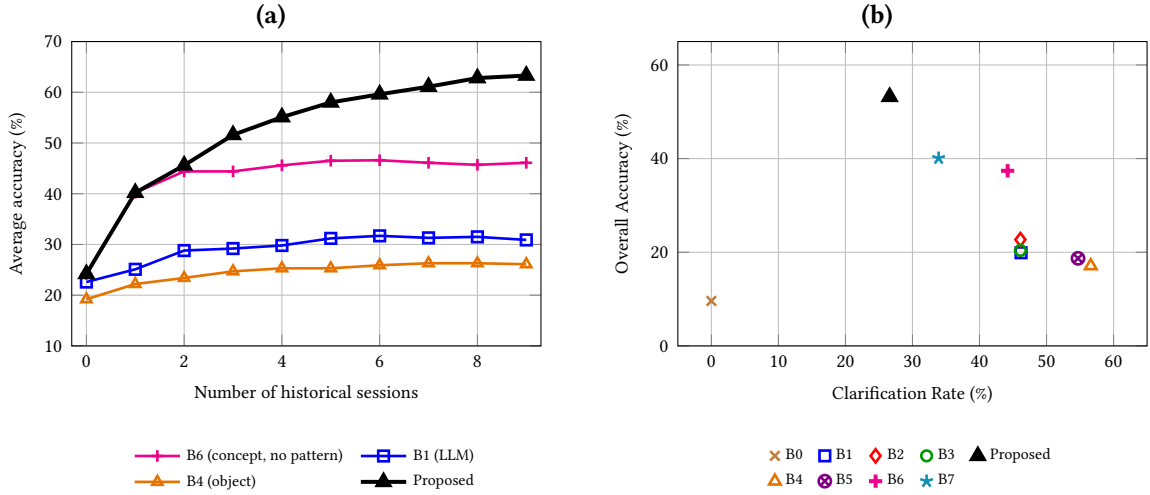

\paragraph{\textbf{Knowledge transfer and impact of user-specific histories (H1-H2).}}
We analyzed performance when test commands were grouped by how often the target object (with ambiguous reference) has appeared in recorded history: rare ($\leq 5$), medium ($6$--$10$), and frequent ($11$--$20$). We also explored how the performance changed as a function of the amount and type of user-specific historical information made available to the AI agent; specifically, we analyzed performance as the number of historical sessions varied from $0$ to $9$, and computed the average accuracy across ambiguity levels. These experiments were conducted for the \textit{noask} mode, with the corresponding results summarized in Table~\ref{tab:freq-transfer},  Figure~\ref{fig:tradeoff-history}(a), and Table~\ref{tab:matched-mismatched}. We observed that:
\begin{itemize}
    \item For rare targets, our architecture nearly tripled the accuracy compared with the baseline that did not include the compositional hierarchy (\textbf{B4}): $44.6\%$ against $16.2\%$, and nearly doubled the accuracy compared with the baseline that does not include the user-specific patterns (\textbf{B6})---see Table~\ref{tab:freq-transfer}. This performance of our architecture is due to the transfer of knowledge (in the form of concepts in the compositional hierarchy) from known (previously seen) objects to the new objects. 
    
    \item Performance with baseline \textbf{B4}, which does not include the compositional hierarchy, remained almost unchanged in as additional historical information was made available---see  Figure~\ref{fig:tradeoff-history}(a)---because object identity does not transfer to novel entities. With our architecture, on the other hand, accuracy improved with access to additional user-specific histories. 
    
    \item User-specific patterns (in terms of hierarchical concepts) make an important contribution to our architecture's performance. Computing the contextual (heuristic) measure ($S_{them}$) for a user based on the stored history of interactions of a different user resulted in performance worse than not using any user-specific history. For example, in Table~\ref{tab:matched-mismatched}, our architecture's accuracy,  averaged over ambiguity levels, dropped from $62.7\%$ to $16.0\%$, which is lower than the accuracy with \textbf{B3}.
\end{itemize}
These findings provide strong support for hypotheses \textbf{H1-H2}.

\paragraph{\textbf{Comparison with the LLM baselines (H2).}}
We then analyzed the performance of our architecture in the context of the LLM baselines (\textbf{B1-B2}) that received the same historical information. Based on results in Table~\ref{tab:bylevel-noask}, Table~\ref{tab:bylevel-ask}  and Figure~\ref{fig:tradeoff-history}, we observe that:
\begin{itemize}
    \item In the \emph{noask} mode, the performance of the LLM baselines was substantially poorer than our architecture, e.g., in Table~\ref{tab:bylevel-noask}, \textbf{B1} and \textbf{B2} achieved $30.9\%$ and $35.9\%$ (respectively) on average against $62.7\%$ with our architecture, and our architecture performed much better at each level of ambiguity. 
    
    \item In the \emph{ask} mode, the LLM baselines requested clarification markedly more often than the proposed method, e.g., in Table~\ref{tab:bylevel-ask}, the AI agent asked for clarification in $\approx 46\%$ of the trials with \textbf{B1-B2} compared with $26.6\%$ with our architecture. Even with these additional clarification queries, the accuracy provided by the LLM baselines was much lower than that of our architecture. This is why the marker for our architecture is closer to the upper left corner of Figure~\ref{fig:tradeoff-history}(b), the ideal point that corresponds to high accuracy and low requirement for clarification, than the markers for the other baselines. 
    
    \item In Figure~\ref{fig:tradeoff-history}(a), we observed that accuracy for the LLM baselines only increased minimally as the AI agent had access to additional sessions of user-specific historical interactions; the accuracy increases but plateaus quickly when user-specific patterns are not modeled and leveraged explicitly. With our architecture, on the other hand, we observed a substantial increase in accuracy as soon as some user-specific information is available because it was able to leverage the user-specific histories to improve performance. 
\end{itemize}
These findings indicate the advantages of the compositional hierarchy and the ability to quickly and accurately model (and leverage) personalized user interactions; they thus strongly support \textbf{H2}.

%%%%%%%%%%%%%%%% ablation by level + concept-vs-pattern %%%%%%%%%%%%%%%%
\begin{figure}[t]
\centering
\begin{minipage}[t]{0.49\textwidth}
\centering
\textbf{(a)}\par\vspace{-5pt}
\begin{tikzpicture}[baseline=(current bounding box.north)]
\begin{axis}[
    width=\linewidth, height=0.72\linewidth,
    xlabel={Ambiguity Level}, ylabel={Overall Accuracy (\%)},
    xmin=0.5, xmax=4.5, ymin=5, ymax=80,
    xtick={1,2,3,4}, xticklabels={A1,A2,A3,A4},
    ytick={10,20,30,40,50,60,70,80},
    grid=major,
    label style={font=\scriptsize}, tick label style={font=\scriptsize},
    legend style={font=\tiny, draw=none, at={(0.5,-0.30)}, anchor=north, legend columns=2},
]
\addplot+[color=orange!90!black, mark=triangle, line width=1pt]
    coordinates {(1,38.7) (2,25.8) (3,19.6) (4,18.7)};
\addlegendentry{B4: Object-level}
\addplot+[color=violet, mark=otimes, line width=1pt]
    coordinates {(1,46.2) (2,28.4) (3,21.8) (4,20.9)};
\addlegendentry{B5: $\mathcal{L}_0$ attributes}
\addplot+[color=blue, mark=square, line width=1pt]
    coordinates {(1,46.5) (2,33.2) (3,23.1) (4,20.9)};
\addlegendentry{B1: LLM+CoT}
\addplot+[color=magenta, mark=+, line width=1.5pt]
    coordinates {(1,63.1) (2,52.0) (3,36.4) (4,35.6)};
\addlegendentry{B6: Concept-level (no pattern)}
\addplot+[color=black, mark=triangle*, line width=1.5pt]
    coordinates {(1,74.2) (2,64.4) (3,57.3) (4,54.7)};
\addlegendentry{Proposed}
\end{axis}
\end{tikzpicture}
\end{minipage}
\hfill
\begin{minipage}[t]{0.49\textwidth}
\centering
\textbf{(b)}\par\vspace{-5pt}
\begin{tikzpicture}[baseline=(current bounding box.north)]
\begin{axis}[
    width=\linewidth, height=0.72\linewidth,
    xlabel={Ambiguity Level}, ylabel={Overall Accuracy (\%)},
    xmin=0.5, xmax=4.5, ymin=30, ymax=80,
    xtick={1,2,3,4}, xticklabels={A1,A2,A3,A4},
    ytick={30,40,50,60,70,80},
    grid=major,
    label style={font=\scriptsize}, tick label style={font=\scriptsize},
    legend style={font=\tiny, draw=none, at={(0.5,-0.30)}, anchor=north, legend columns=1},
]
\addplot+[color=cyan!70!black, mark=o, dashed, line width=1pt]
    coordinates {(1,67.3) (2,54.0) (3,36.4) (4,36.0)};
\addlegendentry{Concept evidence only ($S_{concept}$)}
\addplot+[color=orange!90!black, mark=triangle, dashed, line width=1pt]
    coordinates {(1,61.9) (2,55.7) (3,48.7) (4,48.0)};
\addlegendentry{Pattern evidence only ($S_{pattern}$)}
\end{axis}
\end{tikzpicture}
\end{minipage}
\vspace{-1em}
\caption{\textnormal{(a) Overall accuracy of some baselines and our architecture plotted as a function of ambiguity level in the \emph{noask} mode. The baseline based on object-level representation and the three contextual factors (\textbf{B4}) or that based on atomic attributes (\textbf{B5}) do not perform any better than the LLM baseline (\textbf{B1}), but the baseline that considers the domain-level concepts and heuristically modeled contextual factors (\textbf{B6}) performs substantially better than \textbf{B1}; our architecture, which also considers the user-specific workflow patterns (in $\mathcal{L}_2$) provides a further improvement in performance; (b) Overall accuracy plotted as a function of the ambiguity level in the \emph{noask} mode when ASP-filtered candidates are ranked based on the two heuristic measures that constitute the thematic score ($S_{them}$). Accuracy is higher based on the measure that models concept evidence ($S_{concept}$) at low ambiguity level, with the measure that models pattern evidence ($S_{pattern}$) making a bigger contribution at higher ambiguity levels (A3--A4). } }
\label{fig:ablation-combined}
\end{figure}
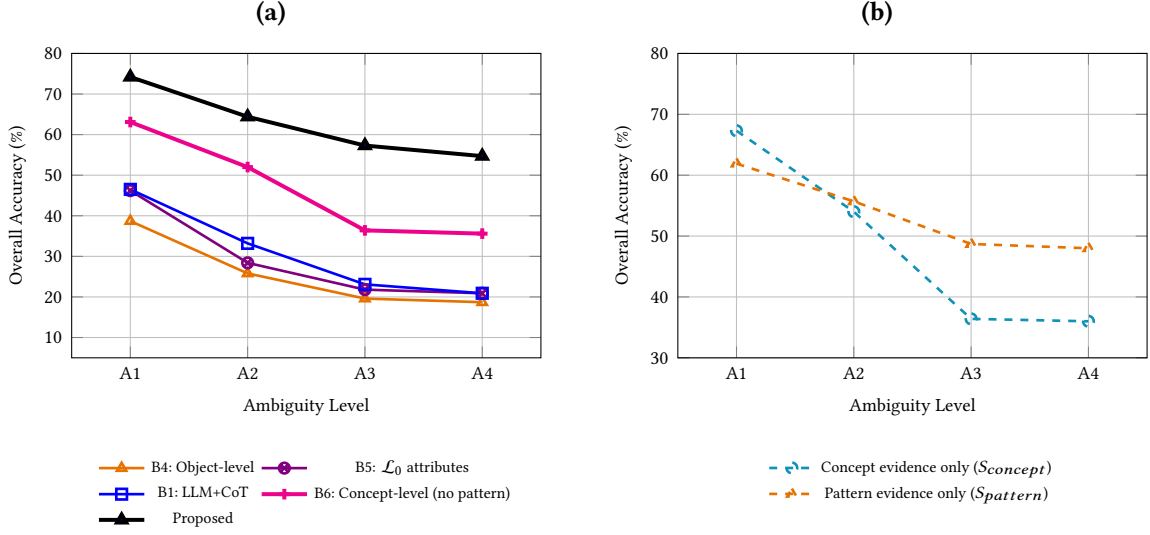

% among the three history representations, object identity (\textbf{B4}) is the weakest because the test targets do not occur in the user's history, whereas concepts transfer through shared attributes, and the full system, which adds workflow patterns, is the most accurate.

\paragraph{\textbf{Impact of compositional hierarchy and contextual factors (H3, H4).}}
Next, we further analyzed the importance of the compositional hierarchy of concepts and the use of heuristically-modeled contextual factors. We did so as a function of the four different ambiguity levels. Based on the results summarized in Table~\ref{tab:bylevel-noask}, Table~\ref{tab:bylevel-ask}, and Figure~\ref{fig:ablation-combined}, we observed the following:
\begin{itemize}
    \item Baseline \textbf{B3} is equivalent to removing the thematic factor ($S_{them}$), which considers the user-specific patterns, from our architecture; this decreased the average accuracy of our architecture from $62.7\%$ to $26.4\%$ in Table~\ref{tab:bylevel-noask}. This is because these patterns provide crucial information, which in conjunction with the semantic similarity and salience scores, enables the AI agent to deal with ambiguities. The results in Figure~\ref{fig:ablation-combined}(a) also support this observation.
    
    \item Using just the object identities with all three contextual factors and ASP-based filtering (\textbf{B4}) resulted in performance somewhat comparable to that of the LLM-based baseline in Figure~\ref{fig:ablation-combined}(a). Also, the overall accuracy reduces substantially as the ambiguity level increases.
    
    \item Using just the atomic attributes level ($\mathcal{L}_0$) along with the contextual factors and ASP-based reasoning (\textbf{B5}) improved performance marginally compared with \textbf{B4}, but performance was still no better than the LLM-based baseline (\textbf{B1}) in Figure~\ref{fig:ablation-combined}(a) since the candidates for disambiguation often had considerable overlap  of the atomic attributes. 
    
    \item Baseline \textbf{B6} corresponds to using the atomic attributes and domain-level concepts, i.e., levels $\mathcal{L}_0$ and $\mathcal{L}_1$ of the compositional hierarchy, along with all three contextual factor and ASP-based reasoning. This option was observed to improve performance considerably compared with \textbf{B4-B5} and the LLM-based baseline (\textbf{B1}) in Figure~\ref{fig:ablation-combined}(a). This is because the use of domain-level concepts supports transfer of existing knowledge to address ambiguity with existing and new objects. Our architecture, which also includes the user-specific patterns of $\mathcal{L}_2$ provides the best performance. 
\end{itemize}
Ranking the baselines based on performance leads to the order: objects $<$ attributes $<$ concepts $<$ concepts$+$patterns, which supports hypotheses \textbf{H3-H4}.

\paragraph{\textbf{Contribution of user-specific workflow patterns (H5).}}
Finally, we further analyzed the importance of acquiring and using the user-specific workflow patterns, and computing and using the thematic factor. The results led to the following observations:
\begin{itemize}
    \item Figure~\ref{fig:ablation-combined}(a) shows the importance of domain-level concepts compared with just using the atomic attributes (i.e., \textbf{B6} compared with \textbf{B5}), particularly when the ambiguity level increases. As stated above, the use of domain-level concepts helps transfer existing knowledge of attribute combination to new objects that share some attributes. As ambiguity increases further, it becomes important to also consider user-specific attribute pattern.
    
    \item Figure~\ref{fig:ablation-combined}(b) summarizes the accuracy when candidates are ranked based on the thematic factor alone, i.e., without the semantic and salience factors. Note that concept evidence ($S_{concept}$ by itself) provides good accuracy ($\approx 67\%$), which is better than the accuracy based on pattern evidence ($S_{pattern}$, $\approx 62\%$) at a low ambiguity level (A1) based on near hypernyms. However, the situation reverses from ambiguity level A2, with user-specific patterns making a strong contribution at ambiguity levels A3--A4. Note that also adding workflow patterns ($\mathcal{L}_2$) raised the accuracy at every ambiguity level, with the largest gains at the higher ambiguity levels A3--A4---see Table~\ref{tab:bylevel-noask} and Figure~\ref{fig:ablation-combined}.
\end{itemize}
These results provide strong evidence in support of \textbf{H5}.

\begin{figure*}[t]
\centering
\definecolor{semcol}{HTML}{4CAF50}
\definecolor{themcol}{HTML}{FF9800}
\definecolor{salcol}{HTML}{1976D2}
\definecolor{filtcol}{HTML}{D9D9D9}
\pgfplotsset{trace/.style={
  ybar stacked, bar width=14pt, scale only axis, width=4.4cm, height=4.2cm,
  ymin=0, ymax=0.62, ytick={0,0.1,0.2,0.3,0.4,0.5,0.6},
  axis x line*=bottom, axis y line*=left, enlarge x limits=0.28,
  title style={font=\small,align=center}, tick label style={font=\small}, label style={font=\small}}}
\begin{minipage}[b]{0.32\textwidth}\centering
\begin{tikzpicture}
\begin{axis}[trace, ylabel={fused score},
  symbolic x coords={lamp,TV,microwave}, xtick=data,
  xticklabels={\strut\textbf{lamp},\strut TV,\strut microwave},
  title={(a) Semantic + thematic\\``Could you switch on \dots''\\(reading context)},
  legend style={at={(0.98,0.98)},anchor=north east,draw=none,fill=none,font=\footnotesize}, legend cell align=left]
\addplot[fill=semcol,draw=none] coordinates {(lamp,0.190)(TV,0.120)(microwave,0.090)};
\addplot[fill=themcol,draw=none] coordinates {(lamp,0.225)(TV,0.1215)(microwave,0.1035)};
\addplot[fill=salcol,draw=none] coordinates {(lamp,0)(TV,0)(microwave,0)};
\legend{semantic,thematic,salience}
\node[anchor=south,font=\small\bfseries] at (axis cs:lamp,0.415){0.42};
\node[anchor=south,font=\small] at (axis cs:TV,0.2415){0.24};
\node[anchor=south,font=\small] at (axis cs:microwave,0.1935){0.19};
\end{axis}
\end{tikzpicture}
\end{minipage}\hfill
\begin{minipage}[b]{0.32\textwidth}\centering
\begin{tikzpicture}
\begin{axis}[trace, ylabel={fused score}, ylabel style={text opacity=0}, yticklabel style={text opacity=0},
  symbolic x coords={banana,apple,orange}, xtick=data,
  xticklabels={\strut\textbf{banana},\strut apple,\strut\textcolor{gray}{orange}},
  title={(b) Thematic decides\\``Please put a fruit on the table.''}]
\addplot[fill=semcol,draw=none] coordinates {(banana,0.196)(apple,0.204)(orange,0)};
\addplot[fill=themcol,draw=none] coordinates {(banana,0.351)(apple,0.099)(orange,0)};
\addplot[fill=salcol,draw=none] coordinates {(banana,0)(apple,0)(orange,0)};
\addplot[fill=filtcol,draw=gray] coordinates {(banana,0)(apple,0)(orange,0.32)};
\node[anchor=south,font=\small\bfseries] at (axis cs:banana,0.547){0.55};
\node[anchor=south,font=\small] at (axis cs:apple,0.303){0.30};
\node[anchor=south,font=\small\itshape,text=gray] at (axis cs:orange,0.32){filtered};
\end{axis}
\end{tikzpicture}
\end{minipage}\hfill
\begin{minipage}[b]{0.32\textwidth}\centering
\begin{tikzpicture}
\begin{axis}[trace, ylabel={fused score}, ylabel style={text opacity=0}, yticklabel style={text opacity=0},
  symbolic x coords={cup,thermos,kettle}, xtick=data,
  xticklabels={\strut\textbf{cup},\strut thermos,\strut kettle},
  title={(c) Salience decides\\``Please give me the hot beverage.''}]
\addplot[fill=semcol,draw=none] coordinates {(cup,0.140)(thermos,0.114)(kettle,0.150)};
\addplot[fill=themcol,draw=none] coordinates {(cup,0.1485)(thermos,0.153)(kettle,0.1485)};
\addplot[fill=salcol,draw=none] coordinates {(cup,0.145)(thermos,0)(kettle,0)};
\node[anchor=south,font=\small\bfseries] at (axis cs:cup,0.4335){0.43};
\node[anchor=south,font=\small] at (axis cs:thermos,0.267){0.27};
\node[anchor=south,font=\small] at (axis cs:kettle,0.2985){0.30};
\end{axis}
\end{tikzpicture}
\end{minipage}
\vspace{-1.5em}
\caption{Fused per-candidate scores for the under-specified commands in each of the three execution traces in Sections~\ref{sec:traces-example1}--\ref{sec:traces-example3}, decomposed into contributions from the three heuristically modeled contextual factors: semantic, salience, and thematic; the total score for the selected candidate is shown in boldface.}
\label{fig:traces}
\end{figure*}
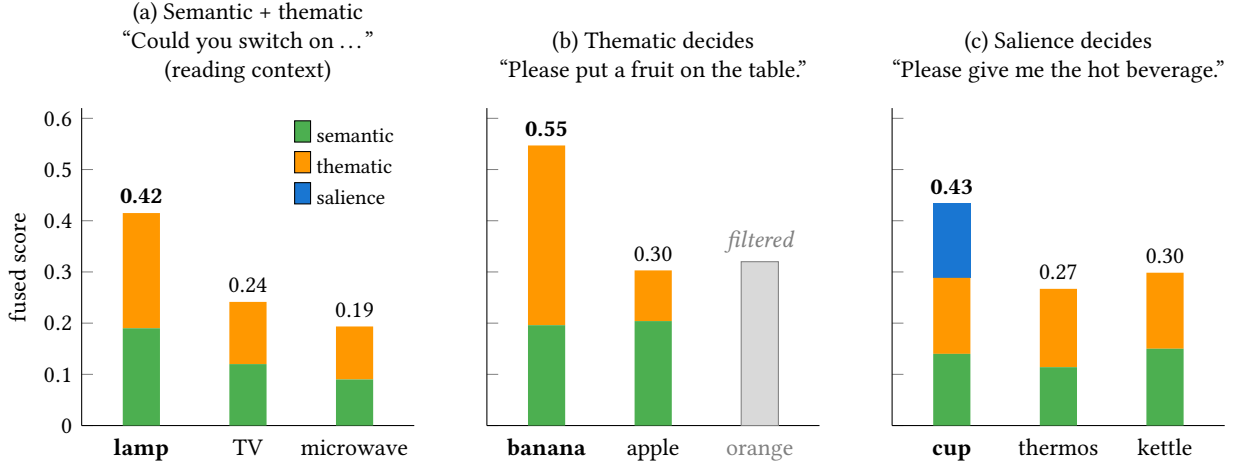
%%%%%%%%%%%

% %%%
% \begin{figure*}[t]
% \centering
% \begin{minipage}[t]{0.49\textwidth}
% \vspace{0pt}\centering
% \includegraphics[width=\linewidth]{log1.png}\\[2ex]
% \includegraphics[width=\linewidth]{log3.png}
% \end{minipage}\hfill
% \begin{minipage}[t]{0.49\textwidth}
% \vspace{0pt}\centering
% \includegraphics[width=\linewidth]{log2.png}
% \end{minipage}
% \caption{Disambiguation scoring traces for Execution Examples 1, 2, and 3 in Section~\ref{sec:traces}.}
% \label{fig:logs}
% \end{figure*}

%%%%%%%%%%%%%%%%%%%%%%%%%%%%%%%%%%%%%%%%%%%%%%%%%%%%%
%%%%%%%%%%%%%%%%%%%%%%%%%%%%%%%%%%%%%%%%%%%%%%%%%%%%%
\section{Execution Traces}
\label{sec:traces}
We present three execution traces to illustrate how the agent combines the semantic compatibility, thematic preference, and session salience signals to resolve under-specified commands. Figure~\ref{fig:traces} reports the fused per-candidate scores for each command. The same three traces are illustrated in the supplementary video.

%%%%%%%%%%%%%%%%%%%%%%%%%%%%%%%%%%%%%%%%%%%%%%%%%%%%%
\subsection{Execution Example 1: Fusion of contextual factors}
\label{sec:traces-example1}
Consider a living room that contains a bookshelf, a table, and three switchable appliances, a lamp, a television, and a microwave, with the agent in the same room. 
\begin{itemize}
    \item Earlier, the user had instructed the AI agent to: \texttt{``Please put the book on the table''}, which the agent executed using the ASP-based plan:
    \begin{align*}
    &occurs(walk(agent1, living\_room), 0), \,\, occurs(walktowards(agent1, bookshelf), 1), \ \\
    &occurs(grab(agent1, book, bookshelf), 2), \,\, occurs(walktowards(agent1, table), 3), \ \\
    &occurs(put(agent1, book, table), 4). 
    \end{align*}
    to achieve \textit{on(book, table)}. The objects changed by this command are recorded as $O^{\Delta}_{t-1}$: \textit{book}, \textit{bookshelf}, and \textit{table}. 
    
    \item The user then speaks to the agent from the other end of the room, and the agent only hears \texttt{``Could you switch on \dots''}. ASP-based feasibility reasoning retains all three appliances \textit{lamp}, \textit{television}, and \textit{microwave} as candidates.
    
    \item The recent action of placing the book on the table is associated with the reading context, identified using semantically relevant data from WordNet, so the \textit{lamp} that shares properties with this context receives the highest semantic compatibility score. 
    
    \item A user-specific workflow pattern mined from this user's history (Section~\ref{sec:problem-compositional}), in which a book-like trigger concept is followed by a lamp-like target concept, gives the \textit{lamp} the highest thematic preference score. 
    
    \item No device was used recently, so session salience does not contribute. The two signals reinforce each other, and the fused score ranks the \textit{lamp} first---see Figure~\ref{fig:traces}(a). As the score for the \textit{lamp} is sufficiently more than that of the second-ranked candidate, the agent commits to \textit{holds(switched\_on(lamp))} as the goal without requesting clarification. The corresponding trace of steps for computing the score for each candidate is shown in Figure~\ref{fig:log1}. 
    
    \item ASP-based reasoning then computes the plan:
    \begin{align*}
    &occurs(walk(agent1, living\_room), 0), \,\, occurs(walktowards(agent1, lamp), 1), \ \\ &occurs(switchon(agent1, lamp), 2).   
    \end{align*}
    which the agent then executes to complete the task. 
\end{itemize}

%%%%%%tracefig1
\vspace{5pt}\hrule height 0.4pt\vspace{3pt}
{\scriptsize
\trow{COMMAND}{}{``Could you switch on \dots''}
\trow{GOAL}{}{switched\_on(agent, switchable\_item)}
\trow{FILTER}{}{switchon accepts only sort \#switch\_furniture, 3 of 66 entities qualify \newline
  candidates = lamp, television, microwave}
\trow{CONTEXT}{}{previous command = ``put the book on the table'', on(book, table) \newline
  changed = book, bookshelf, table \newline
  \sub\begin{tabular}{@{}l@{\hspace{10pt}}l@{}}
  book & reading 0.9, studying 0.5, relaxation 0.1\\
  bookshelf & reading 0.6, storage 0.5, studying 0.2\\
  table & dining 0.8, cooking 0.3, reading 0.1\\
  \end{tabular} \newline
  totals: reading 1.6, dining 0.8, studying 0.7, storage 0.5, cooking 0.3 \newline
  context word = reading}
\trow{SEMANTIC}{semcol}{Similarity to ``reading'' (WordNet) \newline
  \sub lamp 0.475, television 0.300, microwave 0.225 \newline
  the lamp fits a reading situation best}
\trow{THEMATIC}{themcol}{Concept evidence: past switch-on commands \newline
  \sub lamp 0.08, television 0.50, microwave 0.43 \newline
  Pattern evidence: trigger concepts of the previous command \newline
  \sub\begin{tabular}{@{}l@{\hspace{6pt}}l@{}}
  c93 & \{has printing on it, has printed text on the pages, \dots\}\\
  c45 & \{has a flat surface, supports other objects, \dots\}\\
  \end{tabular} \newline
  \sub target concept c21 \{emits light, contains lightbulbs, can be lit for light\}, carried by the lamp \newline
  \sub\begin{tabular}{@{}l@{\hspace{10pt}}l@{\hspace{10pt}}l@{}}
  lamp & c93 to c21 & 0.55\\
       & c45 to c21 & 0.30\\
       & total & 0.85\\
  \end{tabular} \newline
  \sub television, microwave: 0.00 \newline
  combined: lamp 0.50, television 0.27, microwave 0.23 \newline
  the mined pattern outweighs the concept evidence}
\trow{SALIENCE}{salcol}{nothing referenced this session \newline
  \sub lamp 0.00, television 0.00, microwave 0.00}
\trow{SCORE}{}{weights: semantic 0.40, thematic 0.45, salience 0.15 \newline
  \sub\begin{tabular}{@{}l@{\hspace{10pt}}rrr@{\hspace{10pt}}l@{}}
  lamp & 0.475 & 0.50 & 0.00 & = 0.42\\
  television & 0.300 & 0.27 & 0.00 & = 0.24\\
  microwave & 0.225 & 0.23 & 0.00 & = 0.19\\
  \end{tabular}}
\trow{DECIDE}{}{lamp 0.42 over television 0.24, lead 0.75, threshold 0.25 \newline
  committed without asking}
\trow{COMMIT}{}{switched\_on(lamp)}}
\vspace{2pt}\hrule height 0.4pt\vspace{2pt}
\vspace{-1em}
\captionof{figure}{Trace of assigning factor values for each candidate considered for disambiguation in Execution Example 1.}\label{fig:log1}

%%%%%%%%%%%%%%%%%%%%%%%%%%%%%%%%%%%%%%%%%%%%%%%%%%%%%
\subsection{Execution Example 2: Preferences from history of user interactions}
\label{sec:traces-example2}
Next, consider the AI agent assisting a humans in a kitchen, in which an orange is already on the table, and an apple and a banana are within reach on the countertop. 
\begin{itemize}
    \item The human asks the AI agent to: \texttt{``Please put that fruit on the table''}.
    
    \item ASP-based feasibility reasoning first removes the \textit{orange} from further consideration because it is already on the table. This leaves the \textit{apple} and the \textit{banana} as candidates. 
    
    \item The semantic compatibility score is almost the same for these remaining candidates, although there is a slight preference for the \textit{apple}, and neither fruit has been handled recently by this user. 
    
    \item The \textit{banana}, however, shares concepts that recur in this user's records of this action, accumulated in the concept channel $S_{\mathrm{concept}}$, so its thematic preference score is high enough to outweigh the difference in the semantic compatibility score, as shown in Figure~\ref{fig:traces}(b). The corresponding trace of steps executed to compute the scores for each candidate is described in Figure~\ref{fig:log2}. 
    
    \item The agent then computes a plan to achieve the goal of placing the banana on the table, i.e., to achieve \textit{on(banana, table)}. This is indeed the correct solution in this example, but would not have been identified without the explicit modeling and using of the user's history of similar interactions.
\end{itemize}

%%%%%%%%tracefig2
\vspace{5pt}\hrule height 0.4pt\vspace{5pt}
{\scriptsize
\trow{COMMAND}{}{``Please put that fruit on the table.''}
\trow{GOAL}{}{on(fruit, table)}
\trow{FILTER}{}{orange dropped, on(orange, table) already holds \newline
  candidates = banana, apple}
\trow{CONTEXT}{}{previous command = ``put the orange on the table'', changed = orange, table \newline
  context word = dining}
\trow{SEMANTIC}{semcol}{Similarity to ``dining'' (WordNet) \newline
  \sub banana 0.49, apple 0.51 \newline
  near tie, slight edge to the apple}
\trow{THEMATIC}{themcol}{Concept evidence: 98 past put commands \newline
  \sub\begin{tabular}{@{}l@{\hspace{8pt}}l@{\hspace{8pt}}l@{}}
  banana & c34 \{has a soft interior, is soft on the inside, \dots\} & 31 of 98 = 0.32\\
         & c41 \{elongated, is cylindrical with a curved top, \dots\} & 24 of 98 = 0.24\\
         & other concepts & 0.19\\
         & total & 0.75\\
  apple  & c19 \{crisp, crunchy, makes a crunchy sound, \dots\} & 8 of 98 = 0.08\\
         & c23 \{can be red colored, is sweet, is round, \dots\} & 5 of 98 = 0.05\\
         & other concepts & 0.08\\
         & total & 0.21\\
  \end{tabular} \newline
  Pattern evidence: no pattern matched, 0.00 for both \newline
  combined: banana 0.78, apple 0.22 \newline
  this user keeps acting on the concepts the banana carries}
\trow{SALIENCE}{salcol}{nothing referenced this session \newline
  \sub banana 0.00, apple 0.00}
\trow{SCORE}{}{weights: semantic 0.40, thematic 0.45, salience 0.15 \newline
  \sub\begin{tabular}{@{}l@{\hspace{10pt}}rrr@{\hspace{10pt}}l@{}}
  banana & 0.49 & 0.78 & 0.00 & = 0.55\\
  apple & 0.51 & 0.22 & 0.00 & = 0.30\\
  \end{tabular}}
\trow{DECIDE}{}{banana 0.55 over apple 0.30, lead 0.83, threshold 0.25 \newline
  history outweighs the semantic edge of the apple}
\trow{COMMIT}{}{on(banana, table)}}
\vspace{2pt}\hrule height 0.4pt\vspace{2pt}
\vspace{-1em}
\captionof{figure}{Trace of assigning factor values for each candidate considered for disambiguation in Execution  Example 2.}\label{fig:log2}

%%%%%%%%%%%%%%%%%%%%%%%%%%%%%%%%%%%%%%%%%%%%%%%%%%%%%
\subsection{Execution Example 3: Use of salience as a key factor}
\label{sec:traces-example3}
In the third example, consider a kitchen in which a cup, a thermos, and a kettle stand together on the countertop, and the cup has just been filled from the kettle. 
\begin{itemize}
    \item The AI agent is asked by a human to: \texttt{``Please give me the hot beverage''}.

    \item After basic validity checks, the AI agent reasons that all three objects: the \textit{cup}, the \textit{thermos}, and the \textit{kettle}, are feasible candidates.

    \item The user does not show any thematic preference for any of these types of objects, so the semantic compatibility and thematic preference scores are similar for all three candidates. 

    \item Session salience, computed from the maintained world state, assigns the highest weight to the most recently used candidate. Since the \textit{cup} was just filled, it receives the highest (combined) score and is identified as the intended referent---see Figure~\ref{fig:traces}(c) and the trace of steps to compute the scores in Figure~\ref{fig:log3}.

    \item The AI agent then sets its goal to be: \textit{has(user, cup)}, i.e., to ensure that the human user has the target cup. It then computes and executes a plan of actions to achieve this goal.
\end{itemize}

%%%%%%tracefig3
\vspace{5pt}\hrule height 0.4pt\vspace{5pt}
{\scriptsize
\trow{COMMAND}{}{``Please give me the hot beverage.''}
\trow{GOAL}{}{has(user, drinkware)}
\trow{FILTER}{}{nothing dropped, all three can be handed over \newline
  candidates = cup, thermos, kettle}
\trow{CONTEXT}{}{previous command = ``fill the cup from the kettle'', changed = cup \newline
  context word = dining}
\trow{SEMANTIC}{semcol}{Similarity to ``dining'' (WordNet) \newline
  \sub cup 0.35, thermos 0.28, kettle 0.37 \newline
  highest for the kettle, the wrong entity}
\trow{THEMATIC}{themcol}{Concept evidence: past give commands \newline
  \sub cup 0.32, thermos 0.35, kettle 0.33 \newline
  Pattern evidence: no pattern matched, 0.00 for all three \newline
  combined: cup 0.32, thermos 0.35, kettle 0.33 \newline
  same concepts across the three vessels, no separation}
\trow{SALIENCE}{salcol}{recency this session, each step back discounted by 0.9 \newline
  \sub\begin{tabular}{@{}l@{\hspace{10pt}}l@{\hspace{10pt}}l@{}}
  cup & filled one step back & 0.90\\
  thermos & not referenced & 0.00\\
  kettle & not referenced & 0.00\\
  \end{tabular} \newline
  relative: cup 1.00, thermos 0.00, kettle 0.00}
\trow{SCORE}{}{weights: semantic 0.40, thematic 0.45, salience 0.15 \newline
  \sub\begin{tabular}{@{}l@{\hspace{10pt}}rrr@{\hspace{10pt}}l@{}}
  cup & 0.35 & 0.32 & 1.00 & = 0.43\\
  thermos & 0.28 & 0.35 & 0.00 & = 0.27\\
  kettle & 0.37 & 0.33 & 0.00 & = 0.30\\
  \end{tabular}}
\trow{DECIDE}{}{cup 0.43 over kettle 0.30, lead 0.43, threshold 0.25 \newline
  salience alone separates the three}
\trow{COMMIT}{}{has(user, cup)}}
\vspace{2pt}\hrule height 0.4pt\vspace{2pt}
\vspace{-1em}
\captionof{figure}{Trace of assigning factor values for each candidate considered for disambiguation in Execution Example 3.}\label{fig:log3}

\medskip
Jointly, these execution traces demonstrate the working of our architecture and some of its capabilities.

%%%%%%%%%%%%%%%%%%%%%%%%%%%%%%%%%%%%%%%%%%%%%%%%%%%%%%%%%%%%%
%%%%%%%%%%%%%%%%%%%%%%%%%%%%%%%%%%%%%%%%%%%%%%%%%%%%%%%%%%%%%
\section{Conclusions and Future Work}
\label{sec:conclusion}

Foundation models and deep networks are increasingly considered state of the art for AI agents assisting humans, but they find it difficult to provide personalized responses in practical domains characterized by uncertainty and resource limitations. In a departure from methods that focus on the nature and type of training data, computational resources, or optimization methods used in existing monolithic AI models, we explored the design of an architecture for such AI agents based on some fundamental principles, focusing in particular on the principle of hierarchical compositionality and the use of simple heuristics. Also, instead of trying to insert or discover these principles in existing deep networks, we built an architecture based on these principles to better explore the interplay between the different components. As the representative use case for experimental evaluation, we considered an AI agent that had to resolve the ambiguity in the objects being referred to by the human user it is assisting, although disambiguation itself was not the main objective of this work.

\hspace{0.2in} In the design of our architecture, we limited the compositional hierarchy to representing information about domain objects' physical and functional attributes at three levels: (i) atomic attributes drawn from human-validated semantic feature norms; (ii) compound concepts extracted automatically in the form of combinations of these atomic attributes; (iii) combinations of concepts that are extracted automatically to model personalized preferences of specific human users interacting with the agent. We then based the design and use of heuristics on simple rules that advocated for describing objects in terms of the compositional hierarchy, and for resolving the ambiguity in the object being referenced by matching attributes of the candidate objects with those of the known objects that have overlapping attributes and have been used in similar (recent) interactions. This enabled us to design simple heuristic models to estimate and combine semantic similarity, salience, and user-specific thematic preferences. In addition, we enabled the AI agent to perform non-monotonic logical reasoning with current knowledge of domain objects in the compositional hierarchy, some axioms governing domain dynamics, and with the heuristic models of contextual information to achieve the desired disambiguation. As a result, the agent solicits clarification from the human user(s) only when needed.

%\smallskip
\hspace{0.2in} We used ablation studies to experimentally evaluate our architecture against LLM-based baselines and baselines that used various subsets of our architecture's components. The corresponding experimental results showed clear evidence in support of our hypotheses. In particular: 
\begin{enumerate}
    \item integrated reasoning and learning with our compositional hierarchy enables transfer of knowledge from known entities to those that the human user has not interacted with before, leading to more accurate disambiguation and fewer instances of soliciting clarification from a human;
    
    \item our architecture provides a better model of preferences, leading to better performance than the LLM baselines that receive the same information, particularly in scenarios that are more complex and involve personalized interactions; 
    
    \item aggregating statistics to automatically create more abstract representations of objects in the compositional hierarchy yields more accurate disambiguation. Also, the more abstract, user-specific representation is more useful in situations of higher ambiguity; and 
    
    \item the heuristic models for measuring and combining semantic similarity, session salience, and user-specific thematic preference leverage complementary evidence to provide better performance than with models of any individual factor.
\end{enumerate}
Overall, the quantitative and qualitative results demonstrate the clear performance gains obtained over state of the art LLM baselines, and the important contributions made by different components of our architecture based on the underlying principles. 

%\smallskip
\hspace{0.2in} Our architecture opens up multiple directions for further research that we will explore in the future; this includes relaxing the assumptions we have made in the architecture described in this paper. 
\begin{enumerate}
    \item Firstly, we will expand the architecture to consider additional attributes, concepts, and use cases. Specifically, we will explore if and how allowing the revision of all layers of the compositional hierarchy will impact performance. We will also  expand the compositional hierarchy to include temporal concepts, e.g., changes in attributes over time, and a hierarchy of actions (and action sequences) that involve domain objects. This extension will enable us to model, transfer knowledge, and reason about the motion of objects and object parts. 
    
    \item Second, we will explore the design and use of simple heuristics along with models that seek to optimize performance based on all available data. Results documented in this paper and in other work in our lab indicate that simple heuristics are particularly well-suited for making decisions under uncertainty, i.e., when the space of states and outcomes is not well-defined~\cite{dodampegama:FAI26}. Instead of pursuing dual process theories that start with generic commitments about the suitability of specific processes~\cite{kahneman:book11,deneys:book18}, we will continue to match the characteristics of methods with those of the problems to investigate the ecological rationality of different models and methods, i.e., the conditions under which different kinds of models provide good performance. 
    
    \item Third, we would like to investigate the design of neural networks that are based on the principles discussed in this paper. Specifically, unlike the representation and processing choices made in existing networks, our networks would build compositional structures and seek to satisfice under certain conditions in which the optimal choice is not well-defined. 
    
    \item Fourth, we would like to implement and evaluate our architecture on a robot assisting humans in complex indoor domains; this will enable us to study the challenges faced in using such architectures on physical platforms, particularly under conditions of intractability and uncertainty. 
\end{enumerate}
The longer-term goal is to explore the design of architectures that embed and evaluate the impact of different principles, leading to assistive AI agents collaborating robustly with different humans in various complex application domains.

% %
% %% The next line prints the references.
\bigskip
\printbibliography

\end{document}